\documentclass{article}

\usepackage[preprint]{neurips_2024}
\usepackage[pagebackref,breaklinks,colorlinks,allcolors=cvprblue]{hyperref}

\usepackage[utf8]{inputenc} 
\usepackage[T1]{fontenc}    
\usepackage{url}            
\usepackage{booktabs}       
\usepackage{amsfonts}       
\usepackage{nicefrac}       
\usepackage{microtype}      
\usepackage{xcolor}         
\usepackage{wrapfig}
\usepackage{graphicx}
\usepackage{amsmath}
\usepackage{subcaption}
\usepackage{makecell}
\usepackage{algorithm}
\usepackage{algpseudocode}
\usepackage{multirow}
\usepackage{colortbl}

\definecolor{cvprblue}{rgb}{0.21,0.49,0.74}
\definecolor{commentcolor}{RGB}{0, 128, 0} 

\title{FreezeAsGuard: Mitigating Illegal Adaptation of Diffusion Models via Selective Tensor Freezing}

\author{%
	Kai Huang, Haoming Wang and Wei Gao\\
	University of Pittsburgh\\
	\texttt{k.huang, hw.wang, weigao@pitt.edu} \\	 
}

\begin{document}
\maketitle
\begin{abstract}
	Text-to-image diffusion models can be fine-tuned in custom domains to adapt to specific user preferences, but such adaptability has also been utilized for illegal purposes, such as forging public figures' portraits, duplicating copyrighted artworks and generating explicit contents. Existing work focused on detecting the illegally generated contents, but cannot prevent or mitigate illegal adaptations of diffusion models. Other schemes of model unlearning and reinitialization, similarly, cannot prevent users from relearning the knowledge of illegal model adaptation with custom data. In this paper, we present \emph{FreezeAsGuard}, a new technique that addresses these limitations and enables irreversible mitigation of illegal adaptations of diffusion models. Our approach is that the model publisher selectively freezes tensors in pre-trained diffusion models that are critical to illegal model adaptations, to mitigate the fine-tuned model's representation power in illegal adaptations, but minimize the impact on other legal adaptations. 
	Experiment results in multiple text-to-image application domains show that FreezeAsGuard provides 37\% stronger power in mitigating illegal model adaptations compared to competitive baselines, while incurring less than 5\% impact on legal model adaptations.  The source code is available at: \url{https://github.com/pittisl/FreezeAsGuard}.
\end{abstract} 

\vspace{-0.15in}
\section{Introduction}
\vspace{-0.05in}

\begin{wrapfigure}{r}{0.45\textwidth}
	\centering
	\vspace{-0.4in}
	\includegraphics[width=0.45\textwidth]{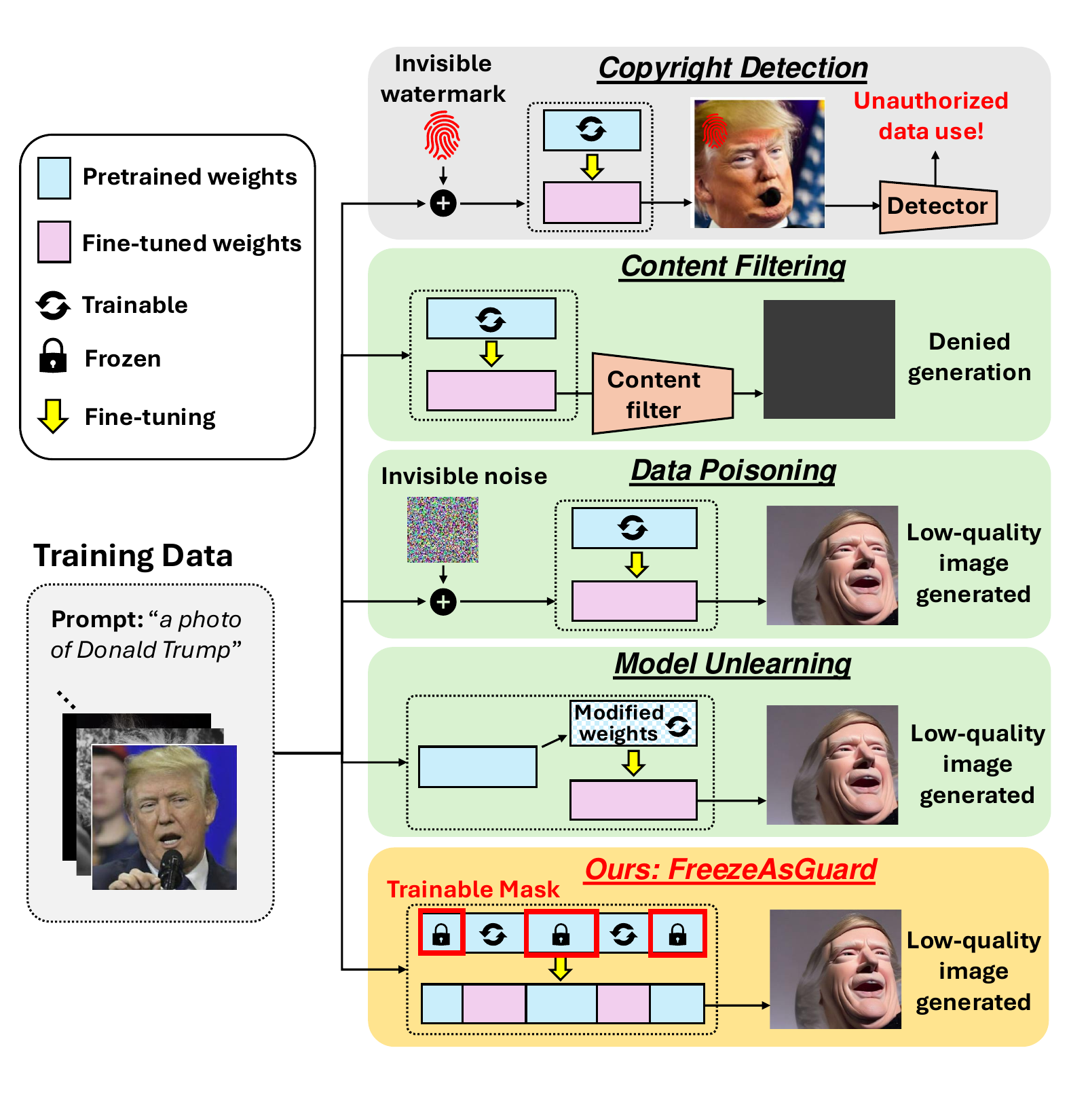}
	\vspace{-0.3in}
	\caption{\label{fig:comparison} Existing work vs. FreezeAsGuard in mitigating malicious adaptation of diffusion models}
	\vspace{-0.25in}
\end{wrapfigure}

Text-to-image diffusion models [\citenum{rombach2022high, podell2023sdxl}] are powerful tools to generate high-quality images aligned with user prompts. After pre-trained by model publishers to embed world knowledge from large image data [\citenum{schuhmann2022laion}], open-sourced diffusion models, such as Stable Diffusion (SD) [\citenum{sd15, sd21}], can be conveniently adapted by users to generate their preferred images\footnote{Many APIs, such as HuggingFace Diffusers [\citenum{wolf2019huggingface}], can be used for fine-tuning open-sourced diffusion models with the minimum user efforts.}, through fine-tuning with custom data in specific domains. For example, diffusion models can be fine-tuned on cartoon datasets to synthesize avatars in video games [\citenum{royer2020xgan}], or on datasets of landscape photos to generate wallpapers [\citenum{diffusionwallpaper}].

\begin{figure*}
	\centering
	\vspace{-0.15in}
	\includegraphics[width=1\linewidth]{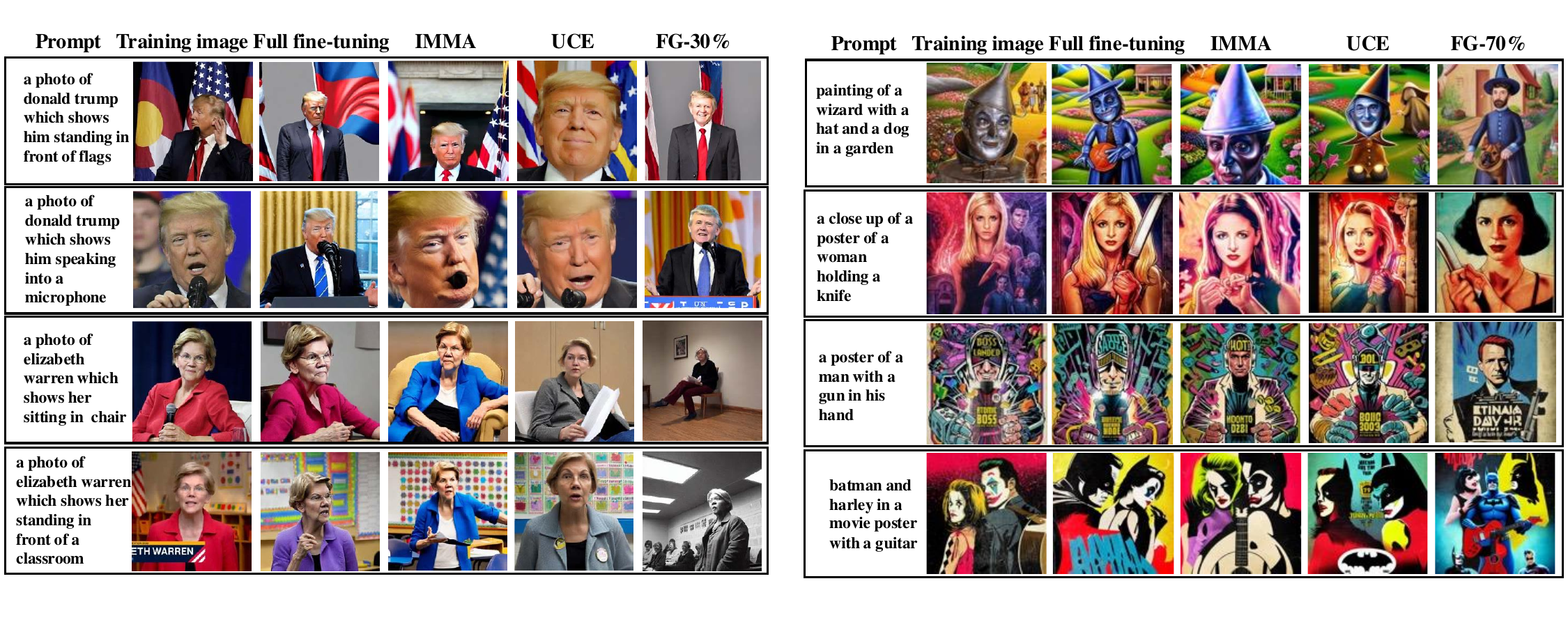}
	\vspace{-0.35in}
	\caption{FreezeAsGuard ensures that portraits (left) and artworks (right) generated by diffusion models in illegal classes cannot be recognizable as target objects, even if the model has been fine-tuned with data samples in illegal classes. In contrast, unlearning schemes (UCE [\citenum{gandikota2024unified}] and IMMA [\citenum{zheng2023imma}]) cannot prevent the unlearned knowledge of illegal classes from being relearned in fine-tuning.}
	\label{fig:illustrative_examples}
\end{figure*}
An increasing risk of democratizing open-sourced diffusion models, however, is that the capability of model adaptation has been utilized for illegal purposes, such as forging public figures' portraits [\citenum{gamage2022deepfakes, gosse2020politics}], duplicating copyrighted artworks [\citenum{heikkila2022artist}], and generating explicit content [\citenum{harwell2017ai}]. Most existing efforts aim to deter attempts of illegal model adaptation with copyright detection [\citenum{zhao2023recipe,cui2023ft,cui2023diffusionshield}], which embeds invisible but detectable watermarks into training data and further generated images, as shown in Figure \ref{fig:comparison}. However, such detection only applies to misuse of training data, and does not mitigate the user's capability of illegal model adaptation. Users can easily bypass such detection by collecting and using their own training data without being watermarked (e.g., users' self-taken photos of public figures).

Instead, an intuitive approach to mitigation is content filtering. However, filtering user prompts [\citenum{derner2023beyond}] can be bypassed by fine-tuning the model to align innocent prompts with illegal image contents [\citenum{webson2021prompt}], and filtering the generated images [\citenum{safetychecker}] is often overpowered with high false-positive rates [\citenum{falsepositive}]. Data poisoning techniques can avoid false positives by injecting invisible perturbations into training data [\citenum{ye2023duaw, zhang2023editguard, shan2023glaze}], but cannot apply when public web data or users' private data is used for fine-tuning. Recent unlearning methods allow model publishers to remove knowledge needed for illegal adaptation by modifying model weights [\citenum{fan2023salun, gandikota2024unified, wu2024erasediff,zheng2023imma}] , but cannot prevent relearning such knowledge via fine-tuning. 



The key limitation of these techniques is that they focus on modifying the training data or model weights, but such modification can be reversed by users via fine-tuning with their own data. Such modification, further, cannot restrain the mitigation power only in \emph{illegal data classes} (e.g., public figures' portraits) without affecting model adaptation in other \emph{legal data classes} (e.g., the user's own portraits), due to the high ambiguity and possible overlap between these classes.

To prevent users from reversing the mitigation maneuvers being applied, in this paper we present \emph{FreezeAsGuard}, a new technique that constrains the trainability of diffusion model's tensors in fine-tuning. As shown in Figure \ref{fig:comparison}, the model publisher selectively freezes tensors in pre-trained models that are critical to fine-tuning in illegal classes (e.g., public figures' portraits), to limit the model's representation power of being fine-tuned in illegal classes. In practice, since most illegal users are not professional and fine-tune diffusion models by simply following the instructions provided by model publishers, tensor freezing can be effectively enforced by model publishers through these instructions, to guide the users to adopt tensor freezing. Essentially, since freezing tensors lowers the trainable model parameters and reduces the computing costs of fine-tuning, users would be well motivated to adopt tensor freezing in fine-tuning practices.

\begin{wrapfigure}{r}{0.25\textwidth}
	\centering
	\vspace{-0.1in}
	\includegraphics[width=0.25\textwidth]{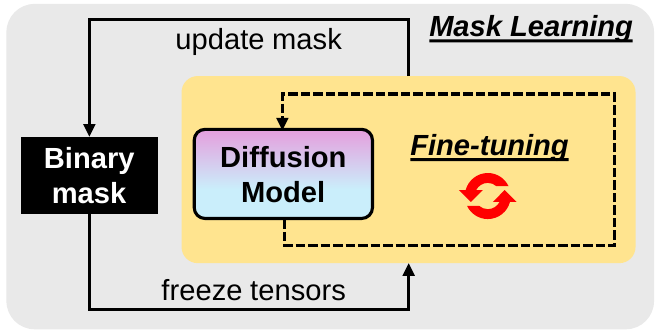}
	\vspace{-0.1in}
	\caption{\label{fig:bilevel_opt} Mask learning and fine-tuning as a bilevel optimization}
	\vspace{-0.1in}
\end{wrapfigure}

The major challenge is how to properly evaluate the importance of tensors in model fine-tuning. Popular attribution-based importance metrics [\citenum{lee2018snip,liu2021group}] are used in model pruning with fixed weight values, but cannot reflect the impact of weight variations in fine-tuning. Such impact of weight variations, in fact, cannot be condensed into a single importance metric, due to the randomness and interdependencies of weight updates in fine-tuning iterations. 

Instead, as shown in Figure \ref{fig:bilevel_opt}, we formulate the selection of frozen tensors in all the illegal classes as one \emph{trainable binary mask}. Given a required ratio of frozen tensors specified by model publisher, we optimize such selection with training data in all the involved illegal classes, through bilevel optimization that combines the iterative process of mask learning and iterations of model fine-tuning. In this way, the mask being trained can timely learn the impact of weight variations on the training loss during fine-tuning. 


With frozen tensors, the model's representation power should be retained when fine-tuned on other legal classes (e.g., user's own portraits). Hence, we  incorporate training samples from legal classes into the bilevel optimization, to provide suppressing signals for selecting tensors being frozen. Hence, the learned mask of freezing tensors should skip tensors that are important to fine-tuning in legal classes. 

We evaluated FreezeAsGuard in three different domains of illegal model adaptations: \emph{1)} forging public figures' portraits, \emph{2)} duplicating copyrighted artworks and \emph{3)} generating explicit contents. For each domain, we use open-sourced or self-collected datasets, and randomly select different data classes as illegal and legal classes. We use competitive model unlearning schemes as baselines, and multiple metrics to measure image quality. Our findings are as follows:
\begin{itemize}
	\item FreezeAsGuard has strong mitigation power in illegal classes. Compared to the competitive baselines, it further reduces the quality of images generated by fine-tuned model by up to 37\%, and ensures the generated images to be unrecognizable as subjects in illegal classes.
	\item FreezeAsGuard has the minimum impact on modal adaptation in legal classes. It ensures on-par quality of the generated images compared to regular full fine-tuning on legal data, with a difference of at most 5\%.
	\item FreezeAsGuard has high compute efficiency. Compared to full fine-tuning, it can save up to 48\% GPU memory and 21\% wall-clock computing time.
\end{itemize}

\vspace{-0.1in}
\section{Background \& Motivation}

\subsection{Fine-Tuning Diffusion Models}
Given text prompts $y$ and images $x$ as training data, fine-tuning a diffusion model approximates the conditional distribution $p(x|y)$ by learning to reconstruct images that are progressively blurred with noise $\epsilon$ over step $t=1,...,T$. Training objective is to minimize the reconstruction loss:
\vspace{-0.05in}
\begin{align}\label{eq:dm_loss}
    \mathcal{L}_{\theta} = \mathbb{E}_{x,y,\epsilon \sim \mathcal{N}(0,1),t}\left[\|\epsilon - \epsilon_{\theta}(\mathcal{E}(x_t), t, \tau(y))\|^2_2\right],
    \vspace{-0.1in}
\end{align}
where $\mathcal{E}(\cdot)$ is the encoder of a pretrained VAE, $\tau(\cdot)$ is a pretrained text encoder, and $\epsilon_{\theta}(\cdot)$ is a denoising model with trainable parameters $\theta$. Most diffusion models adopt UNet architecture [\citenum{ronneberger2015u}] as the denoising model.

In fine-tuning, the diffusion model learns new knowledge by adapting the generic knowledge in the pre-trained model [\citenum{chefer2023hidden}]. For example, new knowledge about ``a green beetle'' can be a combination of generic knowledge on ``hornet'' and ``emerald''. This behavior implies that fine-tuning in different classes may share the same knowledge base, and it is challenging to focus the mitigation power in illegal classes without affecting fine-tuning in other legal classes. This challenge motivates us to regulate FreezeAsGuard's mitigation power by incorporating training samples in legal classes, when selecting tensors being frozen for illegal classes.



\begin{table}[ht]
	\centering
	\vspace{0.15in}
	\fontsize{9}{9}\selectfont
	\begin{tabular}{cccc}
		\toprule
		\makecell{\textbf{Model component} \textbf{Being frozen}} & \textbf{CLIP ($\uparrow$)}	& \textbf{TOPIQ ($\uparrow$)} & \textbf{FID ($\downarrow$)} \\
		\midrule
		No freezing & 31.93 & 0.054 & 202.18 \\
		\midrule
		Attention projectors    & 31.60  & 0.051 & 208.40      \\
		\midrule
		Conv. layers       & 31.54  & 0.047  & 206.58   \\
		\midrule
		Time embeddings       &  31.46 & 0.045 & 212.79     \\
		\midrule
		\makecell{50\% random weights (seed 1)}  & 32.25 & 0.054 & 206.53     \\
		\midrule
		\makecell{50\% random weights (seed 2)} & 32.62 & 0.051 & 216.12 \\			
		\bottomrule
	\end{tabular}
	\vspace{0.1in}
	\captionof{table}{Quality of generated images with different model compoents being frozen,
		using CLIP [\citenum{hessel2021clipscore}], TOPIQ [\citenum{chen2024topiq}], and FID [\citenum{heusel2017gans}] image quality metrics and the captioned pokemon dataset [\citenum{pokemon}]} 
	\vspace{-0.1in}
	\label{tab:different_freezing_examples_numbers}
\end{table}

\begin{wrapfigure}{r}{0.3\textwidth}
	\centering
	\vspace{-0.35in}
	\includegraphics[width=0.3\textwidth]{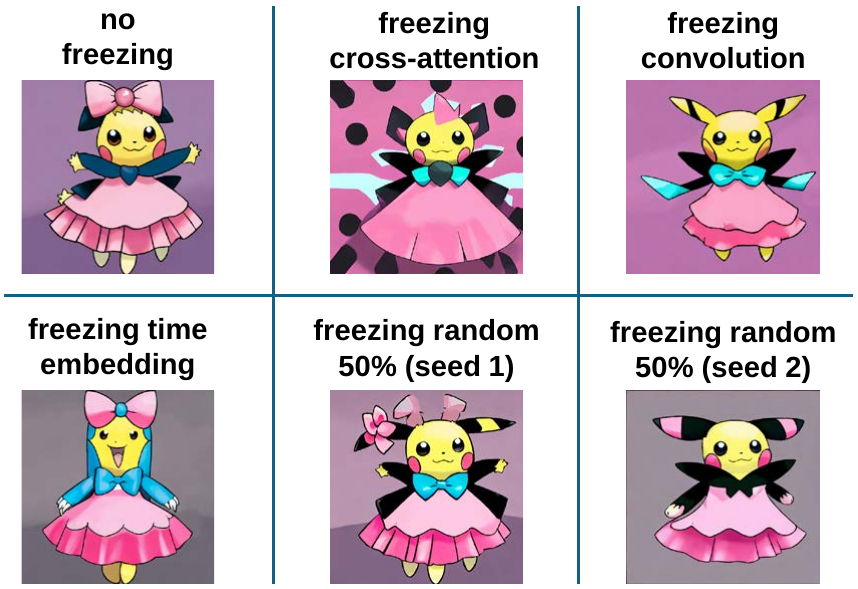}
	\vspace{-0.1in}
	\caption{Generated images with different model components being frozen, with prompt ``a pikachu with a pink dress and a pink bow''}
	\label{fig:different_freezing_examples_images}
	\vspace{-0.15in}
\end{wrapfigure}

\vspace{0.15in}
\subsection{Partial Model Fine-tuning}

An intuitive solution to mitigating illegal model adaptation is to only allow fine-tuning some layers or components of the diffusion model. However, this solution is ineffective in practice, because shallow layers provide primary image features and deep layers enforce domain-specific semantics [\citenum{zeiler2014visualizing}]. They are, hence, both essential to the performance of the fine-tuned models in legal classes.
Similarly, 
as shown in Table \ref{tab:different_freezing_examples_numbers} and Figure \ref{fig:different_freezing_examples_images}, freezing critical model components such as attention projectors and time embeddings can cause large quality drop in generated images. Even when freezing the same amount of model weights (e.g., random 50\%), the exact distribution of frozen weights could also affect the generated images' quality. Such heterogeneity motivates us to instead seek for globally optimal selections of freezing tensors across all model components, by jointly taking all model components into bilevel optimization.

\begin{figure*}[ht]
	\centering
	\vspace{0.1in}
	\includegraphics[width=\linewidth]{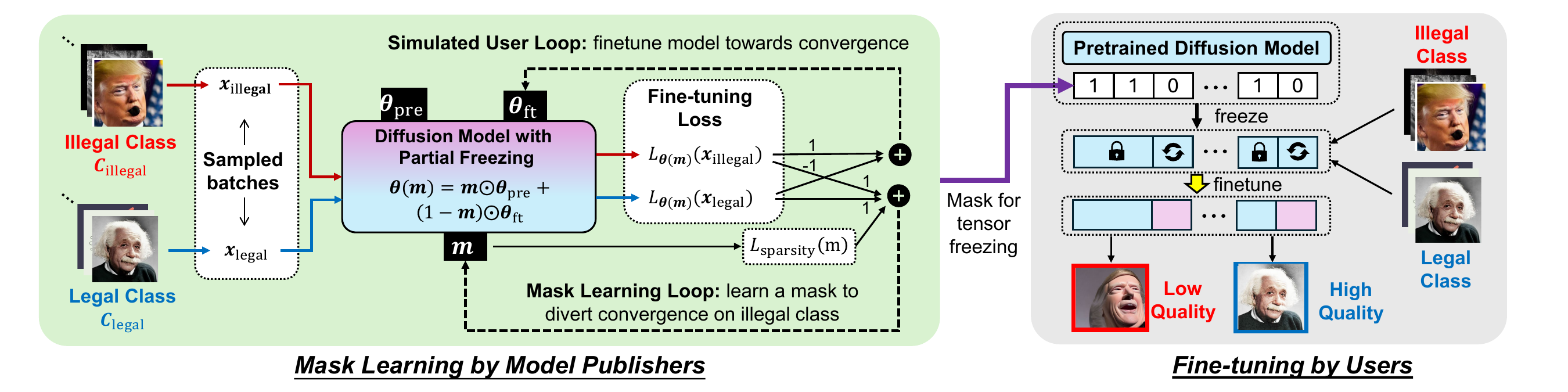}
	\caption{Overview of FreezeAsGuard design}
	\vspace{-0.1in}
	\label{fig:method_overview}
\end{figure*}

\vspace{-0.05in}
\section{Method}
\vspace{-0.05in}	


Our design of FreezeAsGuard builds on bilevel optimization, which embeds one optimization problem within another and both of them are multi-objective optimizations [\citenum{chen2019lambdaopt, liu2018darts,finn2017model}]. This bilevel optimization can be formulated as 
\begin{align}
\mathbf{m}^{*} &= \arg \min_{\mathbf{m}} \left(
-\mathcal{L}_{\boldsymbol{\theta}^{*}(\mathbf{m})}(\mathbf{x}_{illegal}),
\mathcal{L}_{\boldsymbol{\theta}^{*}(\mathbf{m})}(\mathbf{x}_{legal})
\right) \label{eq:upper}
\\
\text{s.t.} & \ \ \ \boldsymbol{\theta}^{*}(\mathbf{m}) = \arg \min_{\boldsymbol{\theta}(\mathbf{m})} \left(
\mathcal{L}_{\boldsymbol{\theta}(\mathbf{m})}(\mathbf{x}_{illegal}),
\mathcal{L}_{\boldsymbol{\theta}(\mathbf{m})}(\mathbf{x}_{legal})
\right), \label{eq:lower}
\end{align}
where $\mathbf{m}$ is the binary mask of selecting frozen tensors, $\mathbf{m}^*$ is the optimized binary mask, $\boldsymbol{\theta}(\mathbf{m})$ represents the model tensors frozen by $\mathbf{m}$, and $\boldsymbol{\theta}^*(\mathbf{m})$ is the converged $\boldsymbol{\theta}(\mathbf{m})$ after fine-tuning. $\mathbf{x}_{illegal}$ and $\mathbf{x}_{legal}$ denote training samples in all the illegal classes ($\mathcal{C}_{illegal}$) and legal classes ($\mathcal{C}_{legal}$), respectively.  
Such bilevel optimization is illustrated in Figure \ref{fig:method_overview}. The lower-level problem in Eq. (\ref{eq:lower}) is a \emph{simulated user loop} that the user fine-tunes the diffusion model by minimizing the loss over both illegal and legal classes. The upper-level problem in Eq. (\ref{eq:upper}) is a \emph{mask learning loop} that learns $\mathbf{m}$ to mitigate the model's representation power when fine-tuned in illegal classes, without affecting fine-tuning in legal classes. 
We use the standard diffusion loss in Eq. (\ref{eq:dm_loss}) and adopt tensor-level freezing to ensure sufficient granularity\footnote{Most existing diffusion models have parameter sizes between 1B and 3.5B, which correspond to at least 686 tensors over the UNet-based denoiser.}, 
without incurring extra computing costs.

To apply the gradient solver, $\mathbf{m}$ and $\boldsymbol{\theta}(\mathbf{m})$ should have differentiable dependencies with the loss function. We model $\boldsymbol{\theta}(\mathbf{m})$ through the weighted summation of pre-trained model tensors $\boldsymbol{\theta}_{pre}$ and fine-tuned model tensors $\boldsymbol{\theta}_{ft}$, such that
\begin{align}\label{eq:frozen_weights}
\boldsymbol{\theta}(\mathbf{m}) = \mathbf{m} \odot \boldsymbol{\theta}_{pre} + (\mathbf{1} - \mathbf{m}) \odot \boldsymbol{\theta}_{ft},
\end{align}
where $\odot$ denotes element-wise multiplication. From the user's perspective, fine-tuning the partially frozen model $\boldsymbol{\theta}(\mathbf{m})$ is equivalent to fine-tuning $\boldsymbol{\theta}_{ft}$, controlled by Eq. (\ref{eq:lower}). To improve compute efficiency, we initialize $\boldsymbol{\theta}_{ft}$ as the fully fine-tuned model tensors on both illegal and legal classes, and gradually enlarge the scope of tensor freezing. Since $\mathbf{m}$ is discrete and not differentiable, we adopt a continuous form $\mathbf{m}(\mathbf{w}) = \sigma(\mathbf{w} / T)$ that applies sigmoid function $\sigma(\cdot)$ over a trainable tensor $\mathbf{w}$. We also did code optimizations for vectorized gradient calculations as in Appendix A.

Note that, although we made $\mathbf{m}$ differentiable in bilevel optimizations, the optimized values in $\mathbf{m}^*$ will be rounded to binary, to ensure complete freezing of selected tensors.



\subsection{Mask Learning in the Upper-level Loop}
To solve the upper-level optimization in Eq. (\ref{eq:upper}), we adopt linear scalarization [\citenum{hwang2012multiple}] to convert it into a single objective $\mathcal{L}_{upper}$ via a weighted summation with weights $(\lambda_1, \lambda_2)$:
\begin{align}\label{eq:loss_upper}
\mathcal{L}_{upper} = -\lambda_1 \mathcal{L}_{\boldsymbol{\theta}^{*}(\mathbf{m})}(\mathbf{x}_{illegal}) + 
\lambda_2 \mathcal{L}_{\boldsymbol{\theta}^{*}(\mathbf{m})}(\mathbf{x}_{legal}),
\end{align}
to involve training samples in both illegal and legal classes when learning $\mathbf{m}$. $(\lambda_1, \lambda_2)$ should ensure that gradient-based feedbacks from the two loss terms are not biased by inequality between the amounts of $\mathbf{x}_{illegal}$ and $\mathbf{x}_{legal}$, and their values should be proportionally set based on these amounts.

Besides, $\mathbf{x}_{illegal}$ and $\mathbf{x}_{legal}$ could contain some knowledge in common, and masked learning from such data may hence affect model adaptation in legal classes. To address this problem, we add a sparsity constraint $\mathcal{L}_{sparsity}$ to $\mathcal{L}_{upper}$ to better control of the mask's mitigation power:
\begin{align}
\mathcal{L}_{sparsity} = \|\mathbf{1}^{\top} \mathbf{m}/N - \rho \|_2^2,
\end{align}
where $N$ is the number of tensors and $\mathbf{1}^{\top} \mathbf{m}/N$ measures the proportion of tensors being frozen. By minimizing $\mathcal{L}_{sparsity}$, the achieved ratio of tensor freezing should approach the given $\rho$. In this way, we can apply gradient descent to minimize $\mathcal{L}_{upper}$ and iteratively refine $\mathbf{m}$ towards optimum.

\subsection{Model Fine-tuning in the Lower-level Loop}
\vspace{-0.05in}
Effectiveness of mask learning at the upper level relies on timely feedback from the lower-level fine-tuning. Every time the mask has been updated by an iteration in the upper level, the lower-level loop should adopt the updated mask into fine-tuning, and return the fine-tuned model tensors and the correspondingly updated loss value as feedback to the upper level. Similar to Eq. (\ref{eq:loss_upper}), the fine-tuning objective is the summation of diffusion losses for illegal and legal domains:
\begin{align}
	\mathcal{L}_{lower} = \mathcal{L}_{\boldsymbol{\theta}^{*}(\mathbf{m})}(\mathbf{x}_{illegal}) + 
	\mathcal{L}_{\boldsymbol{\theta}^{*}(\mathbf{m})}(\mathbf{x}_{legal}).
	\label{eq:loss_lower}
\end{align}

\begin{figure}[ht]
	\centering
	\includegraphics[width=\linewidth]{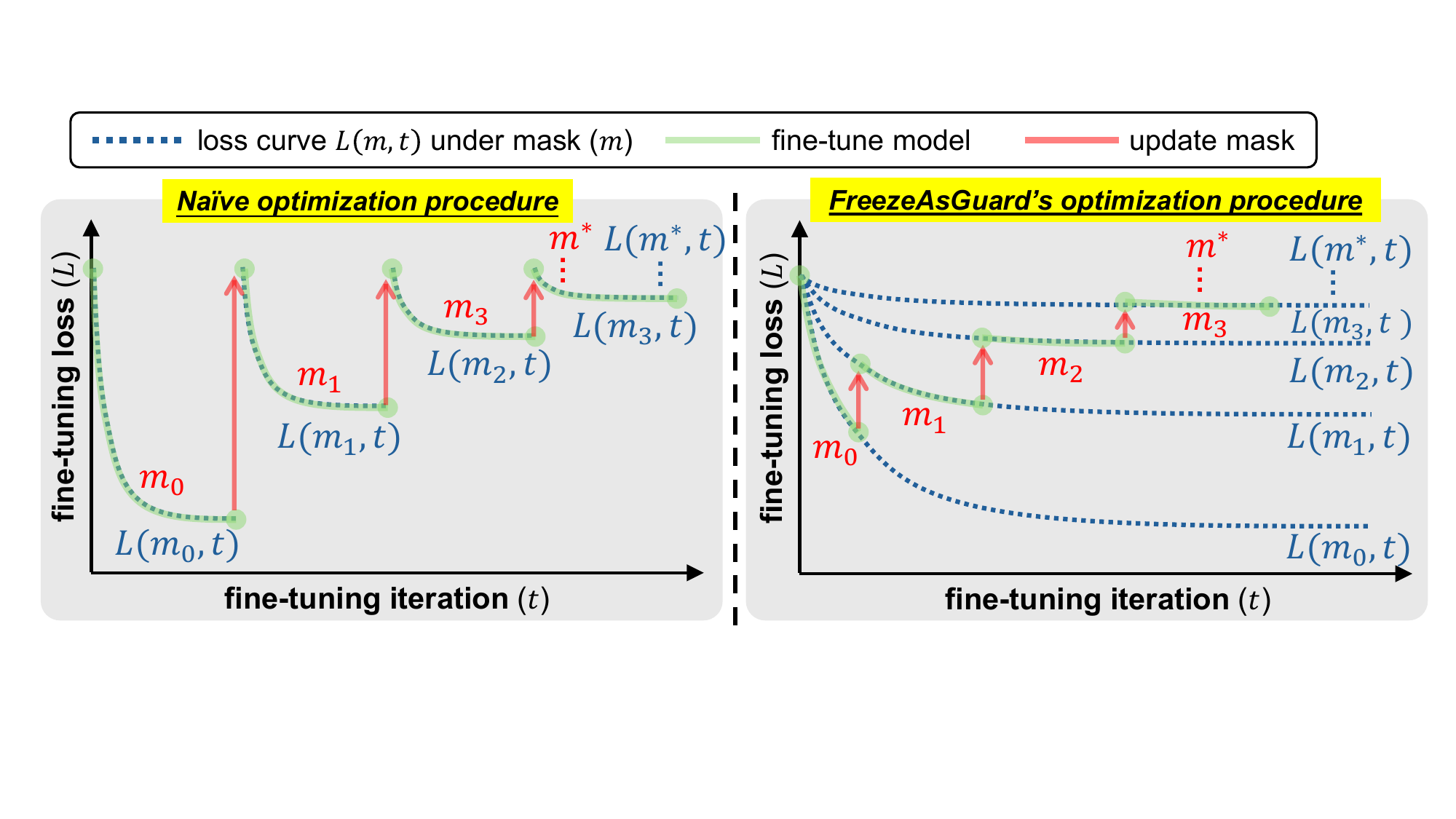}
	\caption{FreezeAsGuard vs. Naive optimization iterations}
	\label{fig:loss_illustration}
	\vspace{-0.1in}
\end{figure}

\subsection{Towards Efficient Bilevel Optimization}
\label{sec:compute_efficiency}
Solving bilevel optimization is computationally expensive, due to the repeated switches between upper-level and lower-level loops [\citenum{ruiz2023dreambooth, zheng2023imma}]. Rigorously, as shown in Figure \ref{fig:loss_illustration} - Left, every time when the mask has been updated, the model should be fine-tuned with a sufficient number of iterations until convergence, before the next update of the mask. 
However, in practice, doing so is extremely expensive. 

Instead, as shown in Figure \ref{fig:loss_illustration} - Right, we observe that the fine-tuning loss typically drops fast in the first few iterations and then violently fluctuates (see Appendix B). Hence, every time in the lower-level loop of model fine-tuning, we do not wait for the loss to converge, but only fine-tune the model for the first few iterations before updating the mask to the upper-level loop of mask learning. After the model update, the fine-tuned model weights are inherited to the next loop of model fine-tuning, to ensure consistency and improve convergence. Hence, the optimization only needs one fine-tuning process, during which the mask can be updated with shorter intervals but higher learning quality. Details of deciding such a number of iterations are in Appendix B.


Further, to perform bilevel optimizations, three versions of diffusion model weights, i.e.,  $\boldsymbol{\theta}(\mathbf{m})$, $\boldsymbol{\theta}_{pre}$ and $\boldsymbol{\theta}_{ft}$, will be maintained for gradient computation. This could significantly increase the memory cost due to large sizes of diffusion models. To reduce such memory cost, we instead maintain only two versions of model weights, namely $\boldsymbol{\theta}(\mathbf{m})$ and $\boldsymbol{\theta}_{d} = \boldsymbol{\theta}_{pre} - \boldsymbol{\theta}_{ft}$. According to Appendix A, the involvement of both $\boldsymbol{\theta}_{pre}$ and $\boldsymbol{\theta}_{ft}$ can be removed by plugging $\boldsymbol{\theta}_{d}$ into the gradient descent calculation. More specifically, for a given model tensor $i$, the gradient descent to update the corresponding mask $m_i$ in the upper-level optimization is:
\begin{equation}
	w_i \leftarrow w_i - \eta_1 \left< \frac{\partial{\mathcal{L}_{upper}}}{\partial{\theta(m)_i}}, \theta_{d}^{(i)} \right> \frac{1}{T}\sigma\left(\frac{w_i}{T}\right)\sigma\left(1-\frac{w_i}{T}\right),
	\label{eq:upper_gd}
\end{equation}
where $\eta_1$ controls the step size of updates and $m_i$ is updated as $\sigma\cdot w_i/{T}$. Further, computing the update of $\boldsymbol{\theta}(\mathbf{m})$ and $\boldsymbol{\theta}_d$ at the lower level should apply the chain rule:
\begin{align}
	\theta_{d}^{(i)} &\leftarrow \theta_{d}^{(i)} + \eta_2 \frac{\partial \mathcal{L}_{lower}}{\partial \theta(m)_{i}} (1-m_i) \label{eq:lower_gd_1}
	\\
	\theta(m)_i &\leftarrow \theta(m)_i - \eta_2 \frac{\partial \mathcal{L}_{lower}}{\partial \theta(m)_{i}} (1-m_i)^2. \label{eq:lower_gd_2}
\end{align}

In this way, as shown in Algorithm \ref{alg:FreezeAsGuard}, FreezeAsGuard alternately runs upper and lower-level gradient descent steps, with the maximum compute efficiency and the minimum memory cost. We initialize the mask to all zeros and $\boldsymbol{\theta}(\mathbf{m})$ starts as a fully fine-tuned model, to mitigate aggressive freezing. In practice, we set random negative values to $\mathbf{w}$ to ensure the continuous form of the mask is near zero.

\vspace{0.05in}
\begin{algorithm}[ht]
	\small
	\caption{Freezing Strategy in FreezeAsGuard}\label{alg:FreezeAsGuard}
	\begin{algorithmic}[1]
		\Require Illegal and legal class data $(\mathcal{C}_{illegal}, \mathcal{C}_{legal})$, step size $\eta_1$ and $\eta_2$, model weights $\boldsymbol{\theta}_{pre}$ and $\boldsymbol{\theta}_{ft}$
		\State $\boldsymbol{\theta}_d \leftarrow \boldsymbol{\theta}_{pre} - \boldsymbol{\theta}_{ft}$, \ \ \ $\mathbf{m} \leftarrow \mathbf{0}$, \ \ \ $\boldsymbol{\theta}(\mathbf{m}) \leftarrow \boldsymbol{\theta}_{ft}$ 
		\For{$k = 1,...,K$}
		
		\For{$l = 1,...,L$}
		\State $(\mathbf{x}_{illegal}, \mathbf{x}_{legal}) \leftarrow$ \texttt{Sample}$(\mathcal{C}_{illegal}, \mathcal{C}_{legal})$
		\State $\frac{\partial{\mathcal{L}_{lower}}}{\partial{\boldsymbol{\theta}(\mathbf{m})}} \leftarrow$ \texttt{Backprop}$(\mathbf{x}_{illegal}, \mathbf{x}_{legal}, \mathcal{L}_{lower}, \boldsymbol{\theta}(\mathbf{m}))$
		\State $(\boldsymbol{\theta}_d, \boldsymbol{\theta}(\mathbf{m})) \leftarrow$ \texttt{Update}$\left(\frac{\partial{\mathcal{L}_{lower}}}{\partial{\boldsymbol{\theta}(\mathbf{m})}}, \mathbf{m}, \boldsymbol{\theta}_d, \boldsymbol{\theta}(\mathbf{m})\right)$ \ \ \ \ \textcolor{commentcolor}{// Refer to Eq. (\ref{eq:lower_gd_1}) and (\ref{eq:lower_gd_2})}
		\EndFor
		
		\State $(\mathbf{x}_{illegal}, \mathbf{x}_{legal}) \leftarrow$ \texttt{Sample}$(\mathcal{C}_{illegal}, \mathcal{C}_{legal})$
		\State $\frac{\partial{\mathcal{L}_{upper}}}{\partial{\boldsymbol{\theta}(\mathbf{m})}} \leftarrow$ \texttt{Backprop}$(\mathbf{x}_{illegal}, \mathbf{x}_{legal}, \mathcal{L}_{upper}, \boldsymbol{\theta}(\mathbf{m}))$
		\State $\mathbf{m} \leftarrow$ \texttt{Update}$\left(\frac{\partial{\mathcal{L}_{upper}}}{\partial{\boldsymbol{\theta}(\mathbf{m})}}, \mathbf{m}, \boldsymbol{\theta}_d, \eta_2\right)$ \ \ \ \ \textcolor{commentcolor}{// Refer to Eq. (\ref{eq:upper_gd})}
		\EndFor \ \ $\Rightarrow$ \textbf{Return} \texttt{Round}$(\mathbf{m})$
	\end{algorithmic}
\end{algorithm}

\section{Experiments}
\label{sec:experiments}
In our experiments, we use three open-source diffusion models, SD v1.4 [\citenum{sd14}], v1.5 [\citenum{sd15}] and v2.1 [\citenum{sd21}], to evaluate three domains of illegal model adaptations: 
\emph{1)} forging public figures' portraits [\citenum{gamage2022deepfakes, gosse2020politics}], \emph{2)} duplicating copyrighted artworks [\citenum{heikkila2022artist}] and \emph{3)} generating explicit content [\citenum{harwell2017ai}]. 

\noindent\textbf{Datasets:} For each domain, we use datasets as listed below, and random select different data classes as illegal and legal classes. We use 50\% of samples in the selected classes for mask learning and model training, and the other samples for testing. More details about datasets are in Appendix C.

\begin{itemize}
	\item \textbf{Portraits of public figures}: We use a self-collected dataset, namely Famous-Figures-25 (FF25), with 8,703 publicly available portraits of 25 public figures on the Web. Each image has a prompt ``a photo of \texttt{<person\_name>} showing \texttt{<content>}'' as description.
	
	\item \textbf{Copyrighted artworks}: We use a self-collected dataset, namely Artwork, which contains 1,134 publicly available artwork images and text captions on the Web, from five famous digital artists with unique art styles. 

	\item \textbf{Explicit contents}: We use the NSFW-caption dataset with 2,000 not-safe-for-work (NSFW) images and their captions [\citenum{explicitdata}] as the illegal class. We use the Modern-Logo-v4 [\citenum{logov4}] dataset, which contains 803 logo images labeled with informative text descriptions, as the legal class.  
	
\end{itemize}

\noindent\textbf{Baseline schemes:} Our baselines include full fine-tuning (FT), random tensor freezing, and two competitive unlearning schemes, namely UCE [\citenum{gandikota2024unified} and IMMA [\citenum{zheng2023imma}]. Existing data poisoning methods [\citenum{ye2023duaw, zhang2023editguard, shan2023glaze}] cannot be used because all data we use is publicly online and cannot be poisoned.

\begin{itemize}
	\item \textbf{Full FT:} It fine-tunes all the tensors of the diffusion model's UNet and has the strongest representation power for adaptation in illegal domains.	
	\item \textbf{Random-$\rho$:} It randomly freezes $\rho$\% of model tensors, as a naive baseline of tensor freezing.
	\item \textbf{UCE [\citenum{gandikota2024unified}]:} It uses unlearning to guide the learned knowledge about illegal classes in the pre-trained model to be irrelevant or more generic. 
	\item \textbf{IMMA [\citenum{zheng2023imma}]:} It reinitializes the model weights so that it is hard for users to conduct effective fine-tuning on the reinitialized model, in both illegal and legal classes. 
\end{itemize}

\noindent\textbf{Measuring image quality:} We used FID [\citenum{heusel2017gans}] and CLIP [\citenum{hessel2021clipscore}] scores to evaluate the quality of generated images. In addition, to better identify domain-specific details in generated images, we also adopted domain-specific image quality metrics, listed as below and described in detail in Appendix D. For each text prompt, the experiment results are averaged from 100 generated images with different random seeds.
\begin{itemize}
	\item \textbf{Domain-specific feature extractors:} Existing work [\citenum{verma2024many}] reported that FID and CLIP fail to measure the similarity between portraits of human subjects, and cannot reflect human perception in images. Hence, for human portraits and artworks, we apply specific feature extractors on real and generated images, and measure the quality of generated images as cosine distance between their feature vectors. For portraits, we use face feature extractors (FN-L, FN, VGG) in DeepFace [\citenum{serengil2024lightface}]. For artworks, we use a pretrained CSD model [\citenum{somepalli2024measuring}]. Details are in Appendix D.1.
	\item \textbf{NudeNet:} We used NudeNet [\citenum{nudenet}] to decide the probability of whether the generated images contain explicit contents, as the image's safety score. Details are in Appendix D.2.
	\item \textbf{Human Evaluation:} To better capture human perception in generated images, we recruited 16 volunteers with diverse backgrounds to provide human evaluations on image quality. For each image, volunteers scored how the generated image is likely to depict the same subject as in the real image from 1 to 7, where 1 means ``very unlikely'' and 7 means ``very likely''. Details are in Appendix D.3.
\end{itemize}


\begin{table}
	\centering
	{\fontsize{9}{9}\selectfont
		\begin{tabular}{ccccccc}
			\toprule
			\multicolumn{2}{c}{\textbf{Metric}} & \textbf{FN-L($\downarrow$)} & \textbf{FN($\downarrow$)} & \textbf{VGG($\downarrow$)} & \textbf{FID($\downarrow$)} & \textbf{Human ($\downarrow$)}\\
			\midrule[1pt]
			\multicolumn{2}{c}{\textbf{Pre-trained model}} & 0.96 &0.92&0.93&164.8& -\\ 			\midrule[1pt]
			\multirow{2}{*}{\textbf{Full FT}}
			&\textbf{illegal} &0.436& 0.455& 0.581& 144.6&6.7 \\ \cmidrule(ll){2-7}
			&\textbf{legal}  &0.436& 0.455& 0.581& 144.6&6.7\\  			\midrule[1pt]
			\multirow{2}{*}{\textbf{UCE}}
			&\textbf{illegal} &0.445 & 0.464 & 0.598 &152.9&4.6  \\ \cmidrule(ll){2-7}
			&\textbf{legal} &0.442& 0.465& 0.583& 151.4&5.4 \\  			\midrule[1pt]
			\multirow{2}{*}{\textbf{IMMA}}
			&\textbf{illegal}&0.467& 0.493& 0.624& 148.8&5.1  \\ \cmidrule(ll){2-7}
			&\textbf{legal}&0.462& 0.475& 0.610& 145.9&5.8  \\  			\midrule[1pt]
			\multirow{2}{*}{\textbf{FG-10\%}}
			&\textbf{illegal}&\textbf{\underline{0.441}}&0.451&0.603& 148.0&\textbf{\underline{4.9}}  \\ \cmidrule(ll){2-7}
			&\textbf{legal}&0.429&0.45& 0.585& 143.6&6.2\\  			\midrule[1pt]
			\multirow{2}{*}{\textbf{R-10\%}}
			&\textbf{illegal}&0.433& 0.451& 0.588& 143.7&6.8  \\ \cmidrule(ll){2-7}
			&\textbf{legal}&0.431&0.457&0.582&144.0 &6.8 \\  			\midrule[1pt]
			\multirow{2}{*}{\textbf{FG-30\%}}
			&\textbf{illegal}&\textbf{\underline{0.482}}& \textbf{\underline{0.504}}&\textbf{\underline{ 0.631}}& \textbf{\underline{153.7}}&\textbf{\underline{3.6}}  \\ \cmidrule(ll){2-7}
			&\textbf{legal} &0.449& 0.478& 0.590& 146.7&6.0 \\  			\midrule[1pt]
			\multirow{2}{*}{\textbf{R-30\%}}
			&\textbf{illegal}&0.429& 0.456& 0.590& 145.0 &5.9 \\ \cmidrule(ll){2-7}
			&\textbf{legal}&0.429& 0.456&0.590&145.0 &5.9 \\  			\midrule[1pt]
			\multirow{2}{*}{\textbf{FG-50\%}}
			&\textbf{illegal}&\textbf{\underline{0.530}}&\textbf{\underline{0.638}}&\textbf{\underline{0.647}}&\textbf{\underline{155.5}}&\textbf{\underline{2.1}}  \\ \cmidrule(ll){2-7}
			&\textbf{legal}&0.499&0.527&0.608&149.5&4.3  \\  			\midrule[1pt]
			\multirow{2}{*}{\textbf{R-50\%}}
			&\textbf{illegal}&0.513& 0.543& 0.638& 151.6 &3.7 \\ \cmidrule(ll){2-7}
			&\textbf{legal}&0.512&0.522&0.632&153.2&3.7  \\  \bottomrule
	\end{tabular}}
	\vspace{0.1in}
	\caption{Mitigation power in 10 illegal classes and 10 legal classes from the FF25 dataset, where worse image quality indicates stronger mitigation power. FG-$\rho$\% means using FreezeAsGuard to freeze $\rho$\% tensors and R-$\rho$\% means random freezing. }
	\vspace{-0.1in}
	\label{tab:ff25_main}
\end{table}

\subsection{Mitigating Forgery of Public Figures' Portraits}
\label{subsec:result_portraits}
We evaluate FreezeAsGuard in mitigating forgery of public figures' portraits, using FF25 dataset and SD v1.5 model. 10 classes are randomly selected from FF25 as illegal and legal classes, respectively. As shown in Table \ref{tab:ff25_main}, FreezeAsGuard can mitigate illegal model adaptation 
by 40\% compared to Full FT. When $\rho$ varies from 10\% to 50\%, it also outperforms the unlearning schemes by 37\%, because these schemes cannot prevent relearning knowledge in illegal classes with new training data. It also ensures better legal model adaptation. With $\rho$=30\%, the impact on legal adaptation is $<$5\%.

\begin{figure}
	\centering
	\vspace{-0.05in}
	\includegraphics[width=0.85\linewidth]{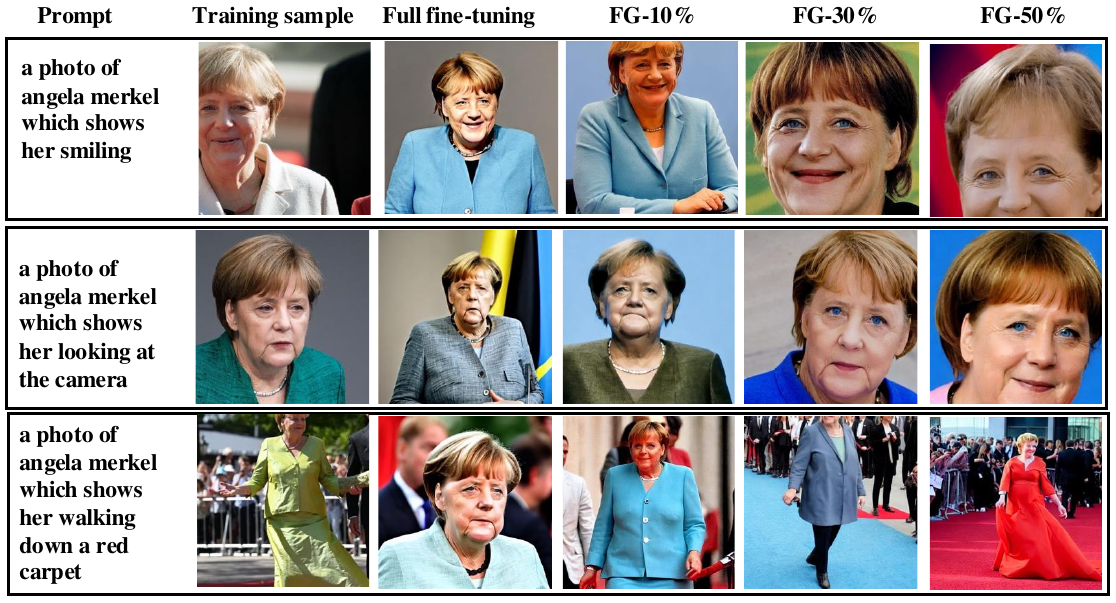}
	\caption{Examples of public figures' portraits generated by FreezeAsGuard under different freezing ratios ($\rho$)}
	\label{fig:ff25_ratio}
	\vspace{-0.25in}
\end{figure}

When the freezing ratio ($\rho$) increases, the difference between FreezeAsGuard and random freezing diminishes, and their mitigation powers also reach a similar level. This means that only a portion of tensors are important for adaptation in specific illegal classes. With a high freezing ratio, random freezing is more likely to freeze these important tensors. Meanwhile, it could also freeze tensors that are important to legal classes, resulting in low performance in legal model adaptations. Hence, as shown in Figure \ref{fig:ff25_ratio}, when $\rho$=30\%, the mitigation power is high enough that the generated images no longer resemble those in training data, and further increasing $\rho$ could largely affect legal model adaptation.

\begin{figure}
	\centering
	\vspace{-0.05in}
	\includegraphics[width=0.9\linewidth]{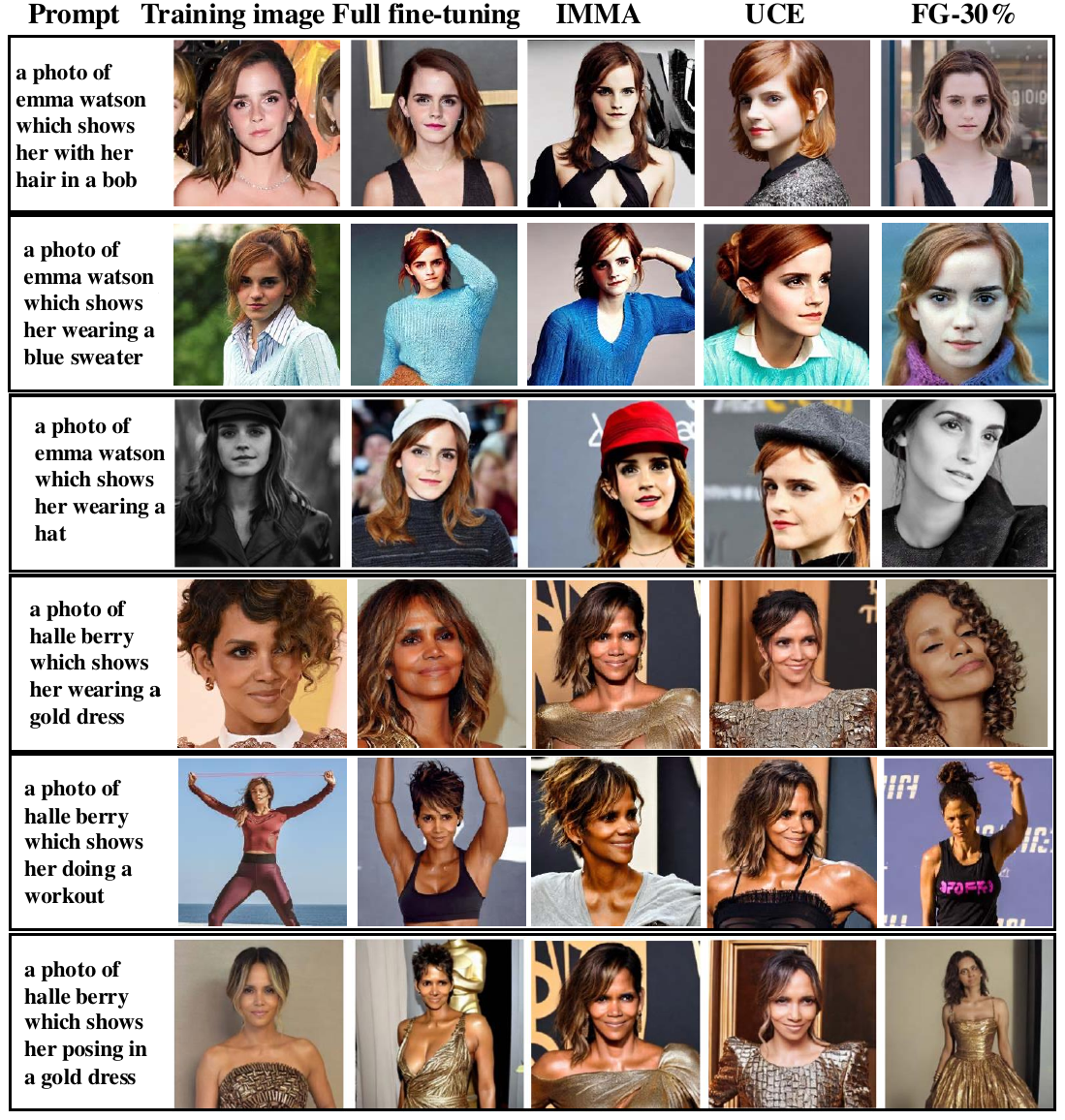}
	\caption{Examples of generated public figures' portraits by FreezeAsGuard with $\rho$=30\% and other baseline methods}
        \label{fig:ff25_example}
\end{figure}

Based on these results, we empirically consider $\rho$=30\% as the optimal freezing ratio on SD v1.5 for the domain of public figures' portraits. Figure \ref{fig:ff25_example} shows example images of baseline methods and FreezeAsGuard with $\rho$=30\%. We can find that FreezeAsGuard effectively prevents the generated images from being recognized as the subjects in illegal classes. Meanwhile, the fine-tuned model can still generate detailed background content and subjects' postures aligned with the prompt, indicating that the mitigation power is highly selective and focuses only on subjects' faces. More examples of generated images are in Appendix F.1.

\begin{table}[ht]
	\centering
	{\fontsize{9}{9}\selectfont
		\begin{tabular}{cccccc}
			\toprule
			\multicolumn{2}{c}{\textbf{Metric}} & \textbf{CSD($\downarrow$)} & \textbf{FID($\downarrow$)} &\textbf{CLIP($\uparrow$)}& \textbf{Human($\downarrow$)}\\
			\midrule[1pt]
			\multicolumn{2}{c}{\textbf{Pre-trained model}} &0.841& 323.8 &-&- \\ \midrule[1pt]
			\multirow{2}{*}{\textbf{Full}}
			&\textbf{illegal}&0.347&187.6&32.31&5.9  \\ \cmidrule(ll){2-6}
			&\textbf{legal}&0.365&194.0&32.19&5.4 \\  \midrule[1pt]
			\multirow{2}{*}{\textbf{UCE}}
			&\textbf{illegal}&0.426& 190.9&32.28&3.3 \\ \cmidrule(ll){2-6}
			&\textbf{legal}&0.381&195.1 &32.17&3.1 \\  \midrule[1pt]
			\multirow{2}{*}{\textbf{IMMA}}
			&\textbf{illegal}&0.396&190.8&32.61&4.6 \\ \cmidrule(ll){2-6}
			&\textbf{legal}&0.377&195&32.98&5.1  \\  \midrule[1pt]
			\multirow{2}{*}{\textbf{FG-30\%}}
			&\textbf{illegal}&\textbf{\underline{0.373}}&\textbf{\underline{190.6}}&32.37&5.7  \\ \cmidrule(ll){2-6}
			&\textbf{legal}&0.382&194.1&32.10&5.2  \\  \midrule[1pt]
			\multirow{2}{*}{\textbf{R-30\%}}
			&\textbf{illegal}&0.351&186.7 &32.45&5.6 \\ \cmidrule(ll){2-6}
			&\textbf{legal}&0.363&194.1&32.56&5.1  \\  \midrule[1pt]
			\multirow{2}{*}{\textbf{FG-50\%}}
			&\textbf{illegal}&\textbf{\underline{0.453}}&\textbf{\underline{194.5}} &\textbf{\underline{32.04}}&\textbf{\underline{3.5}} \\ \cmidrule(ll){2-6}
			&\textbf{legal}&0.40&195.3 &32.49&3.9 \\ \midrule[1pt]
			\multirow{2}{*}{\textbf{R-50\%}}
			&\textbf{illegal}&0.383&189.7 &32.21&5.3 \\ \cmidrule(ll){2-6}
			&\textbf{legal}&0.405&196.0 &32.43&3.7 \\  \midrule[1pt]
			\multirow{2}{*}{\textbf{FG-70\%}}
			&\textbf{illegal}&\textbf{\underline{0.511}}&\textbf{\underline{195.7}}&\textbf{\underline{31.96}}&\textbf{\underline{1.7}} \\ \cmidrule(ll){2-6}
			&\textbf{legal}&0.41&195.3&32.58&3.8  \\  \midrule[1pt]
			\multirow{2}{*}{\textbf{R-70\%}}
			&\textbf{illegal}&0.441&189.2 &32.12&4.9 \\ \cmidrule(ll){2-6}
			&\textbf{legal}&0.454& 196.4 &32.15&4.2 \\  \midrule[1pt]
			\multirow{2}{*}{\textbf{FG-85\%}}
			&\textbf{illegal}&\textbf{\underline{0.574}}& \textbf{\underline{201.2}}&31.74&\textbf{\underline{1.6}}\\ \cmidrule(ll){2-6}
			&\textbf{legal}&0.526& 214.8 &31.91&2.1\\  \midrule[1pt]
			\multirow{2}{*}{\textbf{R-85\%}}
			&\textbf{illegal}&0.565 & 197.6&32.08&2.8\\ \cmidrule(ll){2-6}
			&\textbf{legal}&0.586 & 210.4&32.09&2.7\\ \bottomrule
	\end{tabular}}
	\vspace{0.1in}
	\caption{Mitigation power in one illegal class and one legal class from the Artwork dataset, where worse image quality indicates stronger mitigation power. FG-$\rho$\% means using FreezeAsGuard to freeze $\rho$\% tensors and R-$\rho$\% means random freezing.}
	\vspace{-0.1in}
	\label{tab:main_results}
\end{table}

\begin{figure}[ht]
	\centering
	\vspace{-0.15in}
	\includegraphics[width=0.9\linewidth]{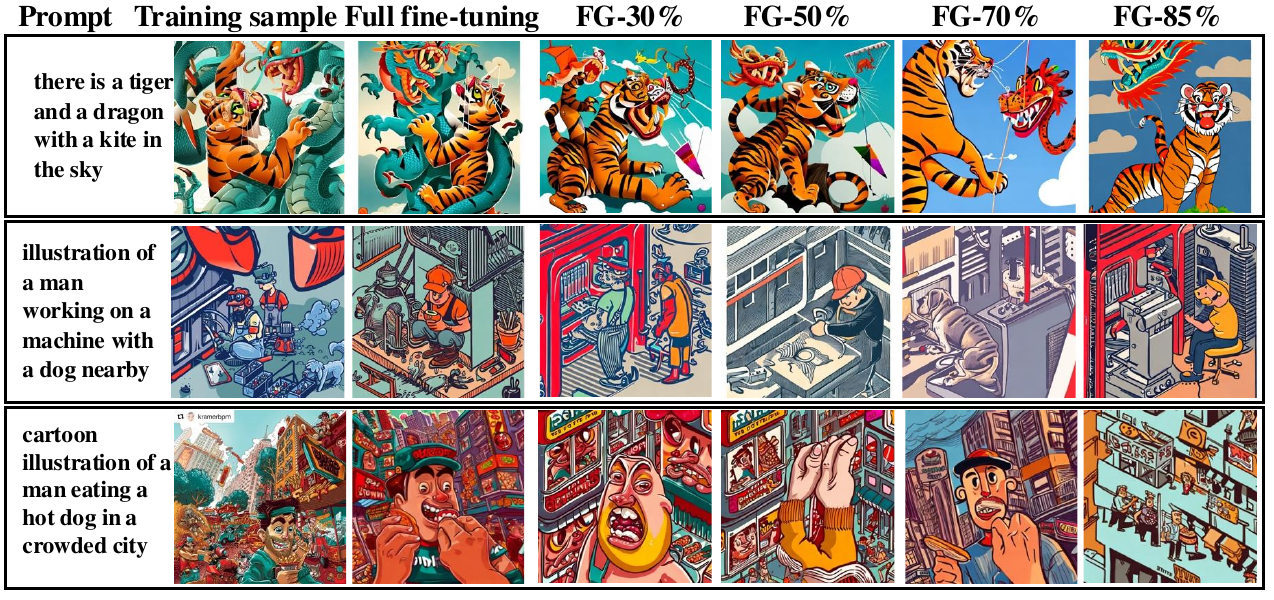}
	\vspace{-0.1in}
	\caption{Examples of artwork images generated by FreezeAsGuard with different freezing ratios}
	\label{fig:art_ratio}
	\vspace{-0.2in}
\end{figure}

\vspace{-0.05in}
\subsection{Mitigating Duplication of Copyright Artworks}
\vspace{-0.05in}
We evaluate the capability of FreezeAsGuard in mitigating the duplication of copyrighted artworks, using the Artwork dataset and SD v2.1 model. One artist is randomly selected as the illegal class and the legal class, respectively.

\begin{figure}[ht]
	\centering
	\vspace{-0.1in}
	\includegraphics[width=0.9\linewidth]{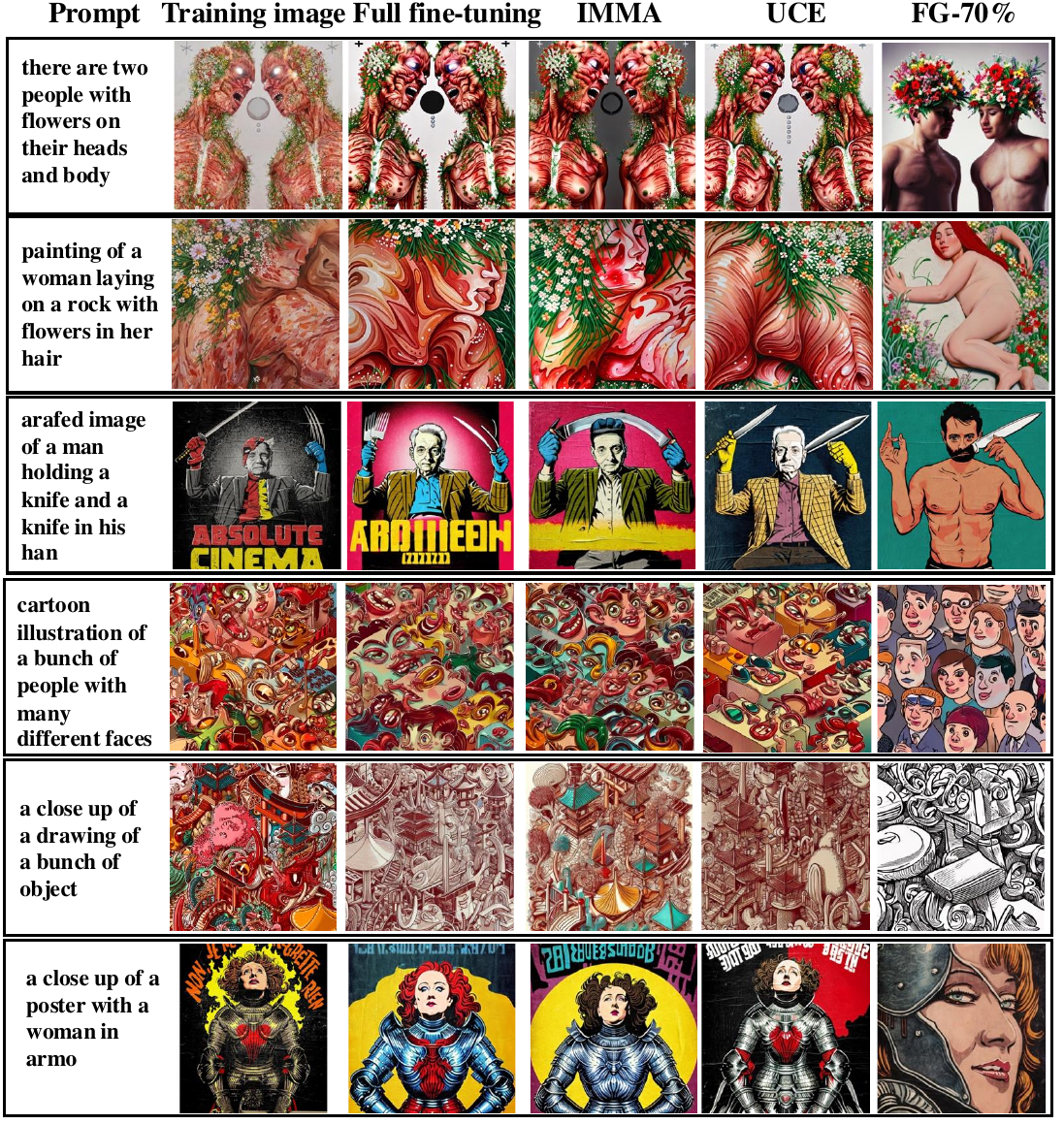}
	\caption{Examples of generated artworks by FreezeAsGuard with $\rho$=70\% and other baseline methods}
        \label{fig:art_example}
	\vspace{-0.1in}
\end{figure}

The results with different freezing ratios are shown in Table \ref{tab:main_results} and Figure \ref{fig:art_ratio}. Unlike results in Section \ref{subsec:result_portraits} where data classes exhibit only subtle differences in facial features, different artists' artworks demonstrate markedly different styles. Hence, a higher freezing ratio is required for sufficient mitigation power. We empirically decide the optimal freezing ratio for the domain of artwork is 70\%. When $\rho$=70\%, FreezeAsGuard can provide 47\% more mitigation power in illegal classes compared to full fine-tuning, and 30\% more compared to unlearning schemes. Figure \ref{fig:art_example} further shows example images generated by FreezeAsGuard with $\rho$=70\%, and more examples can be found in Appendix F.2.

\begin{table}[ht]
	\centering
	{\fontsize{9}{9}\selectfont
		\begin{tabular}{cccc}
			\toprule
			\textbf{Method} &\multicolumn{2}{c}{\textbf{Illegal}} & \textbf{Legal}\\ \cmidrule(ll){2-3}\cmidrule(ll){4-4}
			& NudeNet($\uparrow$) & FID($\downarrow$) & CLIP($\uparrow$)\\ \midrule
			\textbf{Pre-trained model}& 0.47 & - & - \\ \midrule[1pt]
			\textbf{Full FT}& 1.29& 158.1 &32.79 \\ \midrule[1pt]
			\textbf{UCE}&1.20& 158.5& 30.07 \\ \midrule[1pt]
			\textbf{IMMA}&1.17& 162.0& 28.71 \\ \midrule[1pt]
			\textbf{FG-30\%}&\textbf{\underline{1.27}}&\textbf{\underline{159.5}}&32.50\\ \midrule[1pt]
			\textbf{R-30\%}&1.30&158.8&32.79\\ \midrule
			\textbf{FG-50\%}&\textbf{\underline{1.06}}&\textbf{\underline{163.2}}&31.83\\ \midrule[1pt]
			\textbf{R-50\%}&1.20&160.6&30.43\\ \midrule
			\textbf{FG-70\%}&\textbf{\underline{0.87}}&\textbf{\underline{166.1}}&31.56\\ \midrule[1pt]
			\textbf{R-70\%}&1.12&161.8&28.66\\ \midrule
			\textbf{FG-85\%}&\textbf{\underline{0.85}}&\textbf{\underline{166.5}}&30.34\\ \midrule[1pt]
			\textbf{R-85\%}&0.93&164.6&30.81\\ 
			\bottomrule
	\end{tabular}}
	\vspace{0.1in}
	\caption{Mitigation power in illegal class (NSFW-caption dataset) and legal class (Modern-Logo-v4 dataset), where worse image quality (in FID or CLIP) or lower NudeNet score indicates stronger mitigation power. FG-$\rho$\% means using FreezeAsGuard to freeze $\rho$\% tensors and R-$\rho$\% means random freezing.}
	\label{tab:explicit}
\end{table}

\subsection{Mitigating Generation of Explicit Contents}

\begin{figure}
	\centering
	\includegraphics[width=0.8\linewidth]{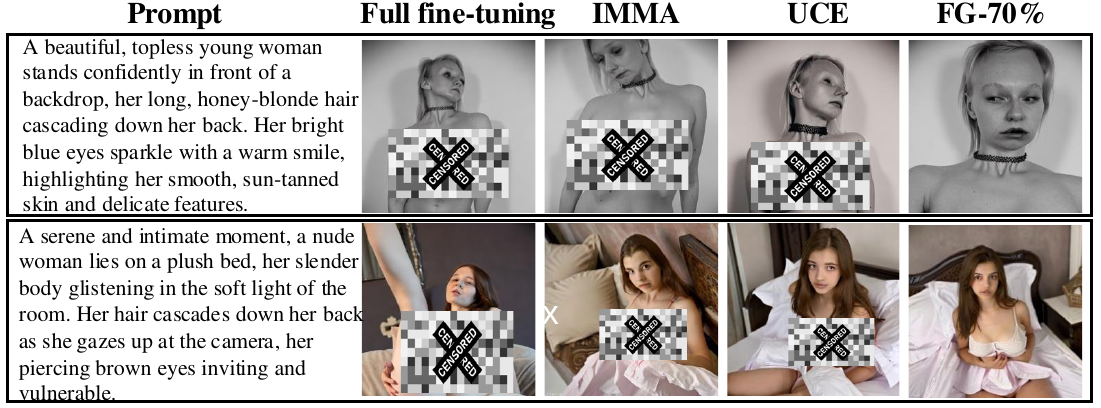}
	\caption{Examples of generated images with explicit contents by FreezeAsGuard with $\rho$=70\% and other baseline methods}
	\label{fig:explicit_example}
	\vspace{-0.1in}
\end{figure}

To evaluate FreezeAsGuard's mitigation of explicit contents, we designate the NSFW-caption dataset as illegal class, 
and the Modern-Logo-v4 dataset as legal class. Results in Table \ref{tab:explicit} and Figure \ref{fig:explicit_example} show that, with $\rho$=70\%, FreezeAsGuard significantly reduces the model's capability of generating explicit contents by up to 38\% compared to unlearning schemes, while maintaining the model's adaptability in legal class. More image examples are in Appendix F.3.

\subsection{Scalability of Mitigation Power}

To evaluate FreezeAsGuard's scalability over multiple illegal classes, we randomly pick 2, 5 and 10 public figures in the FF25 dataset, and  1, 2 and 3 artists in the Artworks dataset, as illegal classes. As shown in Table \ref{tab:ff25_classes} and \ref{tab:art_classes}, when the number of illegal classes increases, FreezeAsGuard can retain strong mitigation power in both cases, and continuously outperforms the unlearning schemes. 
Note that, with more illegal classes, the difference of mitigation power between FreezeAsGuard and random freezing is smaller, because more illegal classes correspond to more adaptation-critical tensors, and random freezing is more likely to cover them.

\begin{table}[ht]
	\centering
	{\fontsize{9}{9}\selectfont
		\begin{tabular}{ccccccc}
			\toprule
			\multirow{2}{*}{\textbf{Method}} & \multicolumn{2}{c}{\textbf{2 classes}} & \multicolumn{2}{c}{\textbf{5 classes}} & \multicolumn{2}{c}{\textbf{10 classes}}\\
			\cmidrule(ll){2-7}
			& illegal &legal& illegal &legal& illegal &legal\\
			\midrule[1pt]
			\textbf{Full FT} &0.397&0.397&0.424&0.424&0.436&0.436\\ \midrule
			\textbf{UCE} &0.435&0.444&0.443&0.437&0.445&0.442\\ \midrule
			\textbf{IMMA} &0.412&0.428&0.461&0.463&0.467&0.462\\ \midrule
			\textbf{FG-30\%} &0.467&0.426&0.474&0.458&0.482&0.449\\
			\bottomrule
	\end{tabular}}
	\vspace{0.1in}
	\caption{Mitigation power in the FF25 dataset, measured by the FN-L score, with different numbers of illegal classes.}
	\vspace{-0.1in}
	\label{tab:ff25_classes}
\end{table}
\vspace{-0.05in}

\begin{table}[ht]
	\centering
	{\fontsize{7.5}{9}\selectfont
		\begin{tabular}{ccccccc}
			\toprule
			\multirow{2}{*}{\textbf{Method}} & \multicolumn{2}{c}{\textbf{1 class}} & \multicolumn{2}{c}{\textbf{2 classes}} & \multicolumn{2}{c}{\textbf{3 classes}}\\
			\cmidrule(ll){2-7}
			& illegal &legal& illegal &legal& illegal &legal\\
			\midrule
			\textbf{Full FT} &0.348& 0.356& 0.415& 0.411& 0.434& 0.458\\ \midrule
			\textbf{UCE} &0.426& 0.381& 0.538& 0.521& 0.552& 0.574 \\ \midrule
			\textbf{IMMA} &0.396 &0.377 &0.483 &0.463 &0.536& 0.496 \\ \midrule
			\textbf{FG-70\%} &0.511 &0.410& 0.609& 0.473& 0.648& 0.525 \\
			\bottomrule
	\end{tabular}}
	\vspace{0.1in}
	\caption{Mitigation power in the Artwork dataset, measured by the CSD score, with different numbers of illegal classes}
	\vspace{-0.2in}
	\label{tab:art_classes}
\end{table}

\subsection{The Learned Selection of Frozen Tensors}
\begin{figure}
	\centering
	\vspace{-0.05in}
	\includegraphics[width=0.7\linewidth]{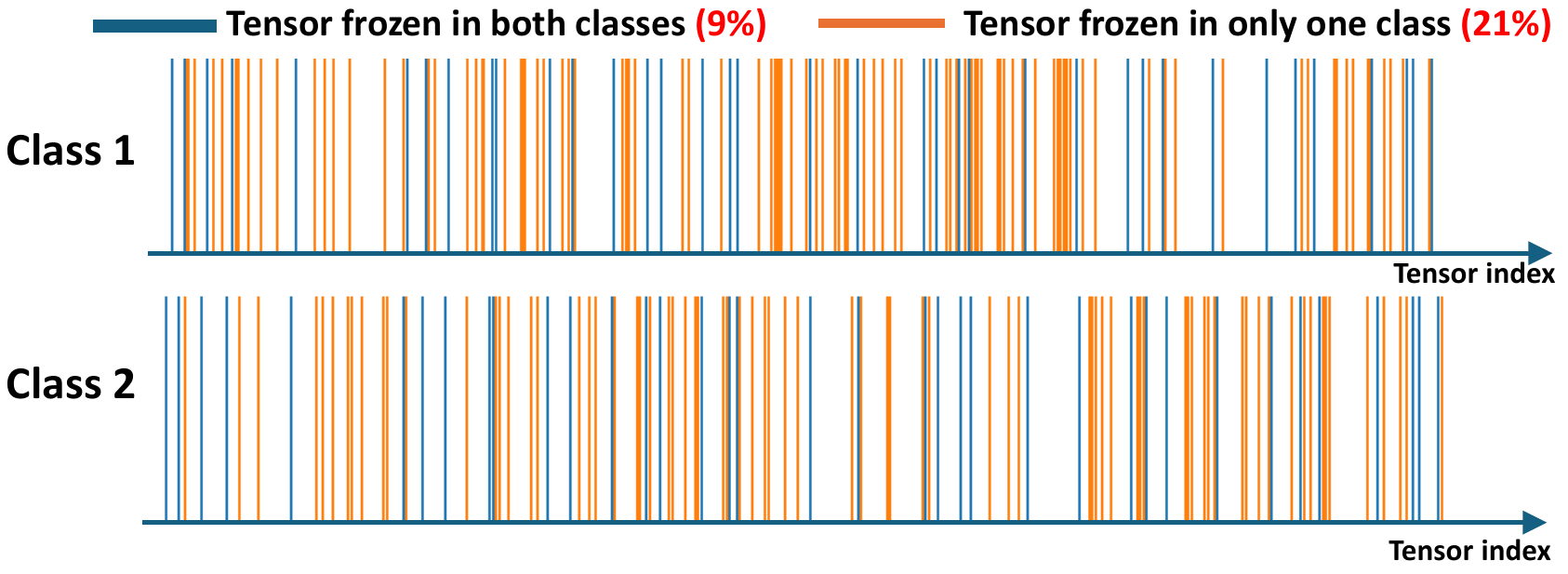}
	\caption{The frozen tensors for illegal classes on the FF-25 dataset, with $\rho$=30\%}
	\label{fig:tensor_ff25}
	\vspace{-0.05in}
\end{figure}

\begin{figure}
	\centering
	\vspace{-0.05in}
	\includegraphics[width=0.7\linewidth]{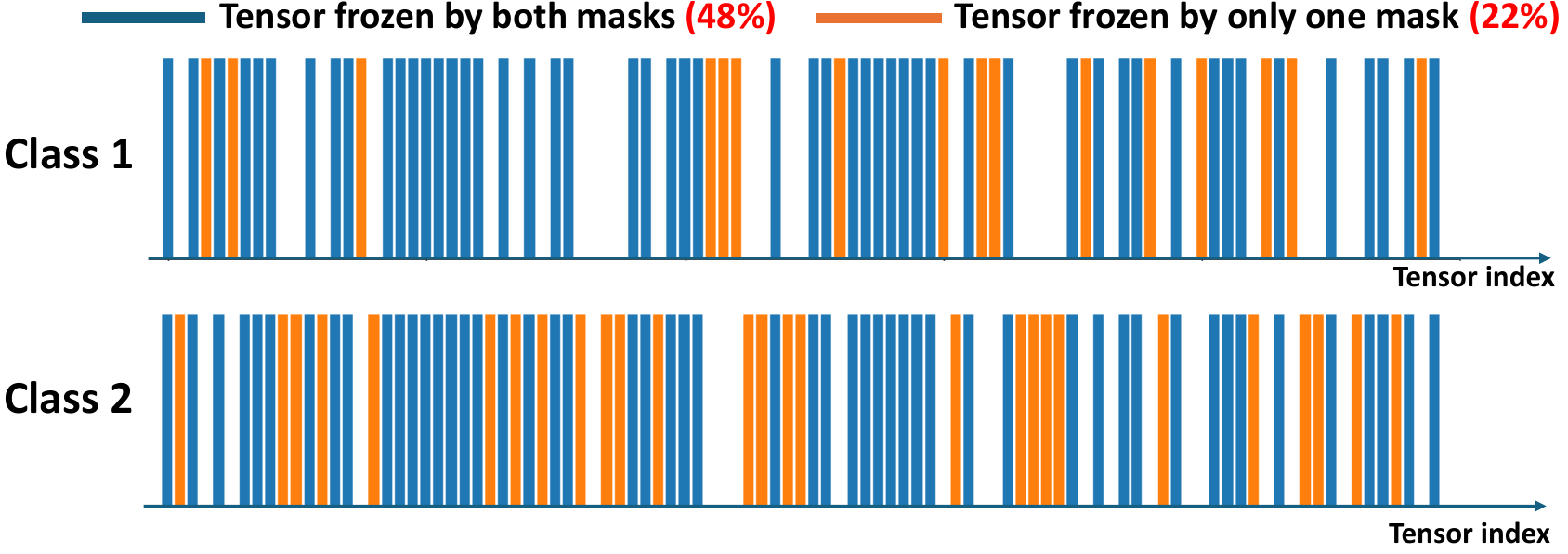}
	\caption{The frozen tensors for illegal classes on the Artwork dataset, with $\rho$=70\%}
	\label{fig:tensor_art}
	\vspace{-0.05in}
\end{figure}

In Figure \ref{fig:tensor_ff25} and \ref{fig:tensor_art}, we visualized the learned binary masks of tensor freezing for different illegal classes on the FF-25 and Artwork datasets, respectively, with the SD v1.5 model. These results show that on both datasets, the tensors being frozen for different illegal classes largely vary, indicating that our mask learning method can properly capture the unique tensors that are critical to each class, hence ensuring scalability. Note that in practice, no matter how many illegal classes are involved, the total amount of frozen tensors will always be constrained by the freezing ratio ($\rho$). When more illegal classes are involved, our results show that FreezeAsGuard is capable of identifying the most critical set of tensors for mitigating the fine-tuned model's representation power.

\subsection{Mitigation Power with Different Models}
As shown in Table \ref{tab:ff25_models}, when applied to different SD models, FreezeAsGuard constantly outperforms baseline schemes. SD v1.4 and v1.5 are generally stronger than SD v2.1, and the gap between illegal and legal classes in FreezeAsGuard is slightly better for v1.4 and v1.5 models. We hypothesize that better pre-trained models have more modularized knowledge distribution over model parameters, and hence allow FreezeAsGuard to have less impact on legal classes.

\begin{table}[ht]
	\centering
	\vspace{-0.05in}
	{\fontsize{9}{9}\selectfont
		\begin{tabular}{ccccccc}
			\toprule
			\multirow{2}{*}{\textbf{Method}} & \multicolumn{2}{c}{\textbf{SD 1.4}} & \multicolumn{2}{c}{\textbf{SD 1.5}} & \multicolumn{2}{c}{\textbf{SD 2.1}}\\
			\cmidrule(ll){2-7}
			& illegal &legal& illegal &legal& illegal &legal\\
			\midrule[1pt]
			\textbf{Full} &0.435&0.435&0.436&0.436&0.439&0.439\\ \midrule
			\textbf{UCE} &0.447&0.442&0.445&0.442&0.445&0.441\\ \midrule
			\textbf{IMMA} &0.451&0.448&0.467&0.462&0.463&0.454\\ \midrule
			\textbf{FG-30\%} &0.489&0.453&0.482&0.449&0.474&0.450\\
			\bottomrule
	\end{tabular}}
	\vspace{0.1in}
	\caption{Mitigation power in the FF25 dataset, measured by the FN-L score, with different diffusion models}
	\vspace{-0.15in}
	\label{tab:ff25_models}
\end{table}
\vspace{-0.05in}

\vspace{-0.05in}
\subsection{Reduction of Computing Costs}
One advantage of freezing tensors is that it reduces the computing costs of fine-tuning.
As shown in Table \ref{tab:fine-tuning_cost}, when fine-tuning the model on a A6000 GPU, by applying FreezeAsGuard's selection of tensor freezing, users can save 22\%-48\% GPU memory and 13\%-21\% wall-clock computing time, compared to other baselines without freezing ($\rho$=0\%). Such savings, hence, well motivate users to adopt the FreezeAsGuard's tensor freezing in their fine-tuning practices.
\begin{table}[ht]
	\centering
	\vspace{-0.05in}
	{\fontsize{9}{9}\selectfont
		\begin{tabular}{ccccc}
			\toprule
			\textbf{Fine-tuning Cost} & $\boldsymbol{\rho}$\textbf{=0\%} & $\boldsymbol{\rho}$\textbf{=1\%} & $\boldsymbol{\rho}$\textbf{=5\%} & $\boldsymbol{\rho}$\textbf{=10\%}  \\
			\midrule
			GPU Memory (GB) & 18.28 & 18.26 & 16.97 & 16.96 \\
			Per-batch computing time (s) & 1.17 & 1.14 & 1.09 & 1.06 \\
			\midrule[1pt]
               \textbf{Fine-tuning Cost} & $\boldsymbol{\rho}$\textbf{=20\%} & $\boldsymbol{\rho}$\textbf{=30\%} & $\boldsymbol{\rho}$\textbf{=40\%} & $\boldsymbol{\rho}$\textbf{=80\%} \\
               \midrule
               GPU Memory (GB)  & 15.43 & 14.15 & 13.61 & 9.49 \\
		    Per-batch computing time (s)  & 1.05 & 1.02 & 1.00 & 0.91 \\
                \bottomrule
	\end{tabular}}
\vspace{0.1in}
\captionof{table}{Computing cost with FreezeAsGuard-$\rho$ on SD v1.5 model, using an NVidia A6000 GPU}
	\label{tab:fine-tuning_cost}
\end{table}

\vspace{-0.05in}
\section{Conclusion \& Broader Impact}
\vspace{-0.05in}
In this paper, we present FreezeAsGuard, a new technique for mitigating illegal adaptation of diffusion models by freezing model tensors that are adaptation-critical only for illegal classes. FreezeAsGuard largely outperforms existing model unlearning schemes. Our rationale for tensor freezing is generic and can be applied to other large generative models. 

\setcitestyle{numbers}
\bibliographystyle{abbrvnat}
\bibliography{ref}

\begin{thebibliography}{65}
\providecommand{\natexlab}[1]{#1}
\providecommand{\url}[1]{\texttt{#1}}
\expandafter\ifx\csname urlstyle\endcsname\relax
  \providecommand{\doi}[1]{doi: #1}\else
  \providecommand{\doi}{doi: \begingroup \urlstyle{rm}\Url}\fi

\bibitem[exp()]{explicitdata}
tungdop2/nsfw\_caption.
\newblock \url{https://huggingface.co/datasets/tungdop2/nsfw_caption}, note =
  {Accessed: 2024-10-30}.

\bibitem[nud()]{nudenet}
Nudenet: lightweight nudity detection.
\newblock \url{https://github.com/notAI-tech/NudeNet}.
\newblock Accessed: 2024-10-30.

\bibitem[fal(2022)]{falsepositive}
some-notes-on-the-stable-diffusion-safety-filter.
\newblock
  \url{https://vickiboykis.com/2022/11/18/some-notes-on-the-stable-diffusion-safety-filter/},
  2022.

\bibitem[aut(2023)]{autocrawler}
Autocrawler.
\newblock \url{https://github.com/YoongiKim/AutoCrawler}, 2023.

\bibitem[log(2023)]{logov4}
modern-logo-v4 dataset.
\newblock
  \url{https://huggingface.co/datasets/logo-wizard/modern-logo-dataset}, 2023.

\bibitem[pok(2023)]{pokemon}
pokemon dataset.
\newblock
  \url{https://huggingface.co/datasets/lambdalabs/pokemon-blip-captions}, 2023.

\bibitem[saf(2023)]{safetychecker}
stable-diffusion-safety-checker.
\newblock \url{https://huggingface.co/CompVis/stable-diffusion-safety-checker},
  2023.

\bibitem[sd1(2023{\natexlab{a}})]{sd14}
stable diffusion v1.4.
\newblock \url{https://huggingface.co/CompVis/stable-diffusion-v1-4},
  2023{\natexlab{a}}.

\bibitem[sd1(2023{\natexlab{b}})]{sd15}
stable diffusion v1.5.
\newblock \url{https://huggingface.co/runwayml/stable-diffusion-v1-5},
  2023{\natexlab{b}}.

\bibitem[sd2(2023)]{sd21}
stable diffusion v2.1.
\newblock \url{https://huggingface.co/runwayml/stable-diffusion-v1-5}, 2023.

\bibitem[dif(2024)]{diffusionwallpaper}
Diffusion wallpaper.
\newblock \url{https://serp.ai/tools/diffusion-wallpaper/}, 2024.

\bibitem[ope(2024)]{opencvfacedet}
Opencv face recognition.
\newblock \url{https://opencv.org/opencv-face-recognition/}, 2024.

\bibitem[Chefer et~al.(2023)Chefer, Lang, Geva, Polosukhin, Shocher, Irani,
  Mosseri, and Wolf]{chefer2023hidden}
H.~Chefer, O.~Lang, M.~Geva, V.~Polosukhin, A.~Shocher, M.~Irani, I.~Mosseri,
  and L.~Wolf.
\newblock The hidden language of diffusion models.
\newblock \emph{arXiv preprint arXiv:2306.00966}, 2023.

\bibitem[Chen et~al.(2024)Chen, Mo, Hou, Wu, Liao, Sun, Yan, and
  Lin]{chen2024topiq}
C.~Chen, J.~Mo, J.~Hou, H.~Wu, L.~Liao, W.~Sun, Q.~Yan, and W.~Lin.
\newblock Topiq: A top-down approach from semantics to distortions for image
  quality assessment.
\newblock \emph{IEEE Transactions on Image Processing}, 2024.

\bibitem[Chen et~al.(2019)Chen, Chen, He, Gao, Li, Lou, and
  Wang]{chen2019lambdaopt}
Y.~Chen, B.~Chen, X.~He, C.~Gao, Y.~Li, J.-G. Lou, and Y.~Wang.
\newblock $\lambda$opt: Learn to regularize recommender models in finer levels.
\newblock In \emph{Proceedings of the 25th ACM SIGKDD International Conference
  on Knowledge Discovery \& Data Mining}, pages 978--986, 2019.

\bibitem[Cui et~al.(2023{\natexlab{a}})Cui, Ren, Lin, Xu, He, Xing, Fan, Liu,
  and Tang]{cui2023ft}
Y.~Cui, J.~Ren, Y.~Lin, H.~Xu, P.~He, Y.~Xing, W.~Fan, H.~Liu, and J.~Tang.
\newblock Ft-shield: A watermark against unauthorized fine-tuning in
  text-to-image diffusion models.
\newblock \emph{arXiv preprint arXiv:2310.02401}, 2023{\natexlab{a}}.

\bibitem[Cui et~al.(2023{\natexlab{b}})Cui, Ren, Xu, He, Liu, Sun, and
  Tang]{cui2023diffusionshield}
Y.~Cui, J.~Ren, H.~Xu, P.~He, H.~Liu, L.~Sun, and J.~Tang.
\newblock Diffusionshield: A watermark for copyright protection against
  generative diffusion models.
\newblock \emph{arXiv preprint arXiv:2306.04642}, 2023{\natexlab{b}}.

\bibitem[Dash et~al.(2022)Dash, Balasubramanian, and
  Sharma]{dash2022evaluating}
S.~Dash, V.~N. Balasubramanian, and A.~Sharma.
\newblock Evaluating and mitigating bias in image classifiers: A causal
  perspective using counterfactuals.
\newblock In \emph{Proceedings of the IEEE/CVF Winter Conference on
  Applications of Computer Vision}, pages 915--924, 2022.

\bibitem[Derner and Batisti{\v{c}}(2023)]{derner2023beyond}
E.~Derner and K.~Batisti{\v{c}}.
\newblock Beyond the safeguards: Exploring the security risks of chatgpt.
\newblock \emph{arXiv preprint arXiv:2305.08005}, 2023.

\bibitem[Fan et~al.(2023)Fan, Liu, Zhang, Wei, Wong, and Liu]{fan2023salun}
C.~Fan, J.~Liu, Y.~Zhang, D.~Wei, E.~Wong, and S.~Liu.
\newblock Salun: Empowering machine unlearning via gradient-based weight
  saliency in both image classification and generation.
\newblock \emph{arXiv preprint arXiv:2310.12508}, 2023.

\bibitem[Finn et~al.(2017)Finn, Abbeel, and Levine]{finn2017model}
C.~Finn, P.~Abbeel, and S.~Levine.
\newblock Model-agnostic meta-learning for fast adaptation of deep networks.
\newblock In \emph{International conference on machine learning}, pages
  1126--1135. PMLR, 2017.

\bibitem[Gamage et~al.(2022)Gamage, Ghasiya, Bonagiri, Whiting, and
  Sasahara]{gamage2022deepfakes}
D.~Gamage, P.~Ghasiya, V.~Bonagiri, M.~E. Whiting, and K.~Sasahara.
\newblock Are deepfakes concerning? analyzing conversations of deepfakes on
  reddit and exploring societal implications.
\newblock In \emph{Proceedings of the 2022 CHI Conference on Human Factors in
  Computing Systems}, pages 1--19, 2022.

\bibitem[Gandikota et~al.(2024)Gandikota, Orgad, Belinkov, Materzy{\'n}ska, and
  Bau]{gandikota2024unified}
R.~Gandikota, H.~Orgad, Y.~Belinkov, J.~Materzy{\'n}ska, and D.~Bau.
\newblock Unified concept editing in diffusion models.
\newblock In \emph{Proceedings of the IEEE/CVF Winter Conference on
  Applications of Computer Vision}, pages 5111--5120, 2024.

\bibitem[Gosse and Burkell(2020)]{gosse2020politics}
C.~Gosse and J.~Burkell.
\newblock Politics and porn: how news media characterizes problems presented by
  deepfakes.
\newblock \emph{Critical Studies in Media Communication}, 37\penalty0
  (5):\penalty0 497--511, 2020.

\bibitem[Harwell(2017)]{harwell2017ai}
D.~Harwell.
\newblock Ai-generated child sex images spawn new nightmare for the web.
\newblock \emph{The Wall Street Journal}, 2017.

\bibitem[Heikkil{\"a}(2022)]{heikkila2022artist}
M.~Heikkil{\"a}.
\newblock This artist is dominating ai-generated art. and he’s not happy
  about it.
\newblock \emph{MIT Technology Review}, 125\penalty0 (6):\penalty0 9--10, 2022.

\bibitem[Hessel et~al.(2021)Hessel, Holtzman, Forbes, Bras, and
  Choi]{hessel2021clipscore}
J.~Hessel, A.~Holtzman, M.~Forbes, R.~L. Bras, and Y.~Choi.
\newblock Clipscore: A reference-free evaluation metric for image captioning.
\newblock \emph{arXiv preprint arXiv:2104.08718}, 2021.

\bibitem[Heusel et~al.(2017)Heusel, Ramsauer, Unterthiner, Nessler, and
  Hochreiter]{heusel2017gans}
M.~Heusel, H.~Ramsauer, T.~Unterthiner, B.~Nessler, and S.~Hochreiter.
\newblock Gans trained by a two time-scale update rule converge to a local nash
  equilibrium.
\newblock \emph{Advances in neural information processing systems}, 30, 2017.

\bibitem[Hwang and Masud(2012)]{hwang2012multiple}
C.-L. Hwang and A.~S.~M. Masud.
\newblock \emph{Multiple objective decision making—methods and applications:
  a state-of-the-art survey}, volume 164.
\newblock Springer Science \& Business Media, 2012.

\bibitem[Jayasumana et~al.(2024)Jayasumana, Ramalingam, Veit, Glasner,
  Chakrabarti, and Kumar]{jayasumana2024rethinking}
S.~Jayasumana, S.~Ramalingam, A.~Veit, D.~Glasner, A.~Chakrabarti, and
  S.~Kumar.
\newblock Rethinking fid: Towards a better evaluation metric for image
  generation.
\newblock In \emph{Proceedings of the IEEE/CVF Conference on Computer Vision
  and Pattern Recognition}, pages 9307--9315, 2024.

\bibitem[Jinjin et~al.(2020)Jinjin, Haoming, Haoyu, Xiaoxing, Ren, and
  Chao]{jinjin2020pipal}
G.~Jinjin, C.~Haoming, C.~Haoyu, Y.~Xiaoxing, J.~S. Ren, and D.~Chao.
\newblock Pipal: a large-scale image quality assessment dataset for perceptual
  image restoration.
\newblock In \emph{Computer Vision--ECCV 2020: 16th European Conference,
  Glasgow, UK, August 23--28, 2020, Proceedings, Part XI 16}, pages 633--651.
  Springer, 2020.

\bibitem[Kahla et~al.(2022)Kahla, Chen, Just, and Jia]{kahla2022label}
M.~Kahla, S.~Chen, H.~A. Just, and R.~Jia.
\newblock Label-only model inversion attacks via boundary repulsion.
\newblock In \emph{Proceedings of the IEEE/CVF conference on computer vision
  and pattern recognition}, pages 15045--15053, 2022.

\bibitem[Karras et~al.(2022)Karras, Aittala, Aila, and
  Laine]{karras2022elucidating}
T.~Karras, M.~Aittala, T.~Aila, and S.~Laine.
\newblock Elucidating the design space of diffusion-based generative models.
\newblock \emph{Advances in Neural Information Processing Systems},
  35:\penalty0 26565--26577, 2022.

\bibitem[Kim and Tompkin(2021)]{kim2021testing}
K.~I. Kim and J.~Tompkin.
\newblock Testing using privileged information by adapting features with
  statistical dependence.
\newblock In \emph{Proceedings of the IEEE/CVF International Conference on
  Computer Vision}, pages 9405--9413, 2021.

\bibitem[Kim et~al.(2022)Kim, Hwang, Ahn, Park, and Kwak]{kim2022learning}
N.~Kim, S.~Hwang, S.~Ahn, J.~Park, and S.~Kwak.
\newblock Learning debiased classifier with biased committee.
\newblock \emph{Advances in Neural Information Processing Systems},
  35:\penalty0 18403--18415, 2022.

\bibitem[Kingma and Ba(2014)]{kingma2014adam}
D.~P. Kingma and J.~Ba.
\newblock Adam: A method for stochastic optimization.
\newblock \emph{arXiv preprint arXiv:1412.6980}, 2014.

\bibitem[Kumar et~al.(2009)Kumar, Berg, Belhumeur, and
  Nayar]{kumar2009attribute}
N.~Kumar, A.~C. Berg, P.~N. Belhumeur, and S.~K. Nayar.
\newblock Attribute and simile classifiers for face verification.
\newblock In \emph{2009 IEEE 12th international conference on computer vision},
  pages 365--372. IEEE, 2009.

\bibitem[Lee et~al.(2018)Lee, Ajanthan, and Torr]{lee2018snip}
N.~Lee, T.~Ajanthan, and P.~H. Torr.
\newblock Snip: Single-shot network pruning based on connection sensitivity.
\newblock \emph{arXiv preprint arXiv:1810.02340}, 2018.

\bibitem[Li et~al.(2023)Li, Li, Savarese, and Hoi]{li2023blip}
J.~Li, D.~Li, S.~Savarese, and S.~Hoi.
\newblock Blip-2: Bootstrapping language-image pre-training with frozen image
  encoders and large language models.
\newblock In \emph{International conference on machine learning}, pages
  19730--19742. PMLR, 2023.

\bibitem[Liu et~al.(2018)Liu, Simonyan, and Yang]{liu2018darts}
H.~Liu, K.~Simonyan, and Y.~Yang.
\newblock Darts: Differentiable architecture search.
\newblock \emph{arXiv preprint arXiv:1806.09055}, 2018.

\bibitem[Liu et~al.(2021)Liu, Zhang, Kuang, Zhou, Xue, Wang, Chen, Yang, Liao,
  and Zhang]{liu2021group}
L.~Liu, S.~Zhang, Z.~Kuang, A.~Zhou, J.-H. Xue, X.~Wang, Y.~Chen, W.~Yang,
  Q.~Liao, and W.~Zhang.
\newblock Group fisher pruning for practical network compression.
\newblock In \emph{International Conference on Machine Learning}, pages
  7021--7032. PMLR, 2021.

\bibitem[Liu et~al.(2015)Liu, Luo, Wang, and Tang]{liu2015faceattributes}
Z.~Liu, P.~Luo, X.~Wang, and X.~Tang.
\newblock Deep learning face attributes in the wild.
\newblock In \emph{Proceedings of International Conference on Computer Vision
  (ICCV)}, December 2015.

\bibitem[Podell et~al.(2023)Podell, English, Lacey, Blattmann, Dockhorn,
  M{\"u}ller, Penna, and Rombach]{podell2023sdxl}
D.~Podell, Z.~English, K.~Lacey, A.~Blattmann, T.~Dockhorn, J.~M{\"u}ller,
  J.~Penna, and R.~Rombach.
\newblock Sdxl: Improving latent diffusion models for high-resolution image
  synthesis.
\newblock \emph{arXiv preprint arXiv:2307.01952}, 2023.

\bibitem[Rombach et~al.(2022)Rombach, Blattmann, Lorenz, Esser, and
  Ommer]{rombach2022high}
R.~Rombach, A.~Blattmann, D.~Lorenz, P.~Esser, and B.~Ommer.
\newblock High-resolution image synthesis with latent diffusion models.
\newblock In \emph{Proceedings of the IEEE/CVF conference on computer vision
  and pattern recognition}, pages 10684--10695, 2022.

\bibitem[Ronneberger et~al.(2015)Ronneberger, Fischer, and
  Brox]{ronneberger2015u}
O.~Ronneberger, P.~Fischer, and T.~Brox.
\newblock U-net: Convolutional networks for biomedical image segmentation.
\newblock In \emph{Medical image computing and computer-assisted
  intervention--MICCAI 2015: 18th international conference, Munich, Germany,
  October 5-9, 2015, proceedings, part III 18}, pages 234--241. Springer, 2015.

\bibitem[Royer et~al.(2020)Royer, Bousmalis, Gouws, Bertsch, Mosseri, Cole, and
  Murphy]{royer2020xgan}
A.~Royer, K.~Bousmalis, S.~Gouws, F.~Bertsch, I.~Mosseri, F.~Cole, and
  K.~Murphy.
\newblock Xgan: Unsupervised image-to-image translation for many-to-many
  mappings.
\newblock \emph{Domain Adaptation for Visual Understanding}, pages 33--49,
  2020.

\bibitem[Ruiz et~al.(2023)Ruiz, Li, Jampani, Pritch, Rubinstein, and
  Aberman]{ruiz2023dreambooth}
N.~Ruiz, Y.~Li, V.~Jampani, Y.~Pritch, M.~Rubinstein, and K.~Aberman.
\newblock Dreambooth: Fine tuning text-to-image diffusion models for
  subject-driven generation.
\newblock In \emph{Proceedings of the IEEE/CVF Conference on Computer Vision
  and Pattern Recognition}, pages 22500--22510, 2023.

\bibitem[Russakovsky et~al.(2015)Russakovsky, Deng, Su, Krause, Satheesh, Ma,
  Huang, Karpathy, Khosla, Bernstein, et~al.]{russakovsky2015imagenet}
O.~Russakovsky, J.~Deng, H.~Su, J.~Krause, S.~Satheesh, S.~Ma, Z.~Huang,
  A.~Karpathy, A.~Khosla, M.~Bernstein, et~al.
\newblock Imagenet large scale visual recognition challenge.
\newblock \emph{International journal of computer vision}, 115:\penalty0
  211--252, 2015.

\bibitem[Schuhmann et~al.(2022)Schuhmann, Beaumont, Vencu, Gordon, Wightman,
  Cherti, Coombes, Katta, Mullis, Wortsman, et~al.]{schuhmann2022laion}
C.~Schuhmann, R.~Beaumont, R.~Vencu, C.~Gordon, R.~Wightman, M.~Cherti,
  T.~Coombes, A.~Katta, C.~Mullis, M.~Wortsman, et~al.
\newblock Laion-5b: An open large-scale dataset for training next generation
  image-text models.
\newblock \emph{Advances in Neural Information Processing Systems},
  35:\penalty0 25278--25294, 2022.

\bibitem[Serengil and Ozpinar(2024)]{serengil2024lightface}
S.~Serengil and A.~Ozpinar.
\newblock A benchmark of facial recognition pipelines and co-usability
  performances of modules.
\newblock \emph{Journal of Information Technologies}, 17\penalty0 (2):\penalty0
  95--107, 2024.
\newblock \doi{10.17671/gazibtd.1399077}.
\newblock URL
  \url{https://dergipark.org.tr/en/pub/gazibtd/issue/84331/1399077}.

\bibitem[Shan et~al.(2023)Shan, Cryan, Wenger, Zheng, Hanocka, and
  Zhao]{shan2023glaze}
S.~Shan, J.~Cryan, E.~Wenger, H.~Zheng, R.~Hanocka, and B.~Y. Zhao.
\newblock Glaze: Protecting artists from style mimicry by $\{$Text-to-Image$\}$
  models.
\newblock In \emph{32nd USENIX Security Symposium (USENIX Security 23)}, pages
  2187--2204, 2023.

\bibitem[Somepalli et~al.(2024)Somepalli, Gupta, Gupta, Palta, Goldblum,
  Geiping, Shrivastava, and Goldstein]{somepalli2024measuring}
G.~Somepalli, A.~Gupta, K.~Gupta, S.~Palta, M.~Goldblum, J.~Geiping,
  A.~Shrivastava, and T.~Goldstein.
\newblock Measuring style similarity in diffusion models.
\newblock \emph{arXiv preprint arXiv:2404.01292}, 2024.

\bibitem[Tan et~al.(2019)Tan, Chan, Aguirre, and Tanaka]{artgan2018}
W.~R. Tan, C.~S. Chan, H.~Aguirre, and K.~Tanaka.
\newblock Improved artgan for conditional synthesis of natural image and
  artwork.
\newblock \emph{IEEE Transactions on Image Processing}, 28\penalty0
  (1):\penalty0 394--409, 2019.
\newblock \doi{10.1109/TIP.2018.2866698}.
\newblock URL \url{https://doi.org/10.1109/TIP.2018.2866698}.

\bibitem[Verma et~al.(2024)Verma, Rassin, Das, Bhatt, Seshadri, Shah, Bilmes,
  Hajishirzi, and Elazar]{verma2024many}
S.~Verma, R.~Rassin, A.~Das, G.~Bhatt, P.~Seshadri, C.~Shah, J.~Bilmes,
  H.~Hajishirzi, and Y.~Elazar.
\newblock How many van goghs does it take to van gogh? finding the imitation
  threshold.
\newblock \emph{arXiv preprint arXiv:2410.15002}, 2024.

\bibitem[Webson and Pavlick(2021)]{webson2021prompt}
A.~Webson and E.~Pavlick.
\newblock Do prompt-based models really understand the meaning of their
  prompts?
\newblock \emph{arXiv preprint arXiv:2109.01247}, 2021.

\bibitem[Wolf et~al.(2019)Wolf, Debut, Sanh, Chaumond, Delangue, Moi, Cistac,
  Rault, Louf, Funtowicz, et~al.]{wolf2019huggingface}
T.~Wolf, L.~Debut, V.~Sanh, J.~Chaumond, C.~Delangue, A.~Moi, P.~Cistac,
  T.~Rault, R.~Louf, M.~Funtowicz, et~al.
\newblock Huggingface's transformers: State-of-the-art natural language
  processing.
\newblock \emph{arXiv preprint arXiv:1910.03771}, 2019.

\bibitem[Wu et~al.(2024)Wu, Le, Hayat, and Harandi]{wu2024erasediff}
J.~Wu, T.~Le, M.~Hayat, and M.~Harandi.
\newblock Erasediff: Erasing data influence in diffusion models.
\newblock \emph{arXiv preprint arXiv:2401.05779}, 2024.

\bibitem[Xu et~al.(2023)Xu, Long, and Nie]{xu2023learning}
W.~Xu, C.~Long, and Y.~Nie.
\newblock Learning dynamic style kernels for artistic style transfer.
\newblock In \emph{Proceedings of the IEEE/CVF Conference on Computer Vision
  and Pattern Recognition}, pages 10083--10092, 2023.

\bibitem[Ye et~al.(2023)Ye, Huang, An, and Wang]{ye2023duaw}
X.~Ye, H.~Huang, J.~An, and Y.~Wang.
\newblock Duaw: Data-free universal adversarial watermark against stable
  diffusion customization.
\newblock \emph{arXiv preprint arXiv:2308.09889}, 2023.

\bibitem[Yu et~al.(2023)Yu, Yu, Yu, Huang, and Li]{yu2023language}
L.~Yu, B.~Yu, H.~Yu, F.~Huang, and Y.~Li.
\newblock Language models are super mario: Absorbing abilities from homologous
  models as a free lunch.
\newblock \emph{arXiv preprint arXiv:2311.03099}, 2023.

\bibitem[Zeiler and Fergus(2014)]{zeiler2014visualizing}
M.~D. Zeiler and R.~Fergus.
\newblock Visualizing and understanding convolutional networks.
\newblock In \emph{Computer Vision--ECCV 2014: 13th European Conference,
  Zurich, Switzerland, September 6-12, 2014, Proceedings, Part I 13}, pages
  818--833. Springer, 2014.

\bibitem[Zhang et~al.(2023)Zhang, Li, Yu, Xu, Li, and
  Zhang]{zhang2023editguard}
X.~Zhang, R.~Li, J.~Yu, Y.~Xu, W.~Li, and J.~Zhang.
\newblock Editguard: Versatile image watermarking for tamper localization and
  copyright protection.
\newblock \emph{arXiv preprint arXiv:2312.08883}, 2023.

\bibitem[Zhang et~al.(2020)Zhang, Deng, Wang, Hu, Li, Zhao, and
  Wen]{zhang2020global}
Y.~Zhang, W.~Deng, M.~Wang, J.~Hu, X.~Li, D.~Zhao, and D.~Wen.
\newblock Global-local gcn: Large-scale label noise cleansing for face
  recognition.
\newblock In \emph{Proceedings of the IEEE/CVF Conference on Computer Vision
  and Pattern Recognition}, pages 7731--7740, 2020.

\bibitem[Zhao et~al.(2023)Zhao, Pang, Du, Yang, Cheung, and
  Lin]{zhao2023recipe}
Y.~Zhao, T.~Pang, C.~Du, X.~Yang, N.-M. Cheung, and M.~Lin.
\newblock A recipe for watermarking diffusion models.
\newblock \emph{arXiv preprint arXiv:2303.10137}, 2023.

\bibitem[Zheng and Yeh(2023)]{zheng2023imma}
Y.~Zheng and R.~A. Yeh.
\newblock Imma: Immunizing text-to-image models against malicious adaptation.
\newblock \emph{arXiv preprint arXiv:2311.18815}, 2023.

\end{thebibliography}


\newpage
\appendix

\section{Vectorizing the Gradient Calculations in Bilevel Optimization}
\label{sec:vectorizing}
In practice, the solutions to bilevel optimization in Eq. (2) and Eq. (3) can usually be approximated through gradient-based optimizers. However, existing deep learning APIs (e.g., TensorFlow and PyTorch) maintain model tensors in either list or dictionary-like structures, and hence the gradient calculation for Eq. (4) cannot be automatically vectorized with the mask vector $\mathbf{m}$. To enhance the compute efficiency, we decompose the process of gradient calculation and assign the majority of compute workload to the highly optimized APIs. 

Specifically, in mask learning in the upper-level loop specified in Eq. (5), $\mathcal{L}_{upper}$'s gradient w.r.t a model tensor's $w_i$ can be decomposed via the chain rule as:
\begin{align}\label{eq:bp_upper}
\frac{\partial{\mathcal{L}_{upper}}}{\partial{w_i}} & =
\left< \frac{\partial{\mathcal{L}_{upper}}}{\partial{\theta(m)_i}}, 
\frac{\partial{\theta(m)_i}}{m_i} \right>
\frac{\partial m_i}{\partial w_i} \\
&= \left< \frac{\partial{\mathcal{L}_{upper}}}{\partial{\theta(m)_i}},
\theta_{pre}^{(i)} - \theta_{ft}^{(i)} \right> 
\frac{1}{T}\sigma\left(\frac{w_i}{T}\right)\sigma\left(1-\frac{w_i}{T}\right),
\end{align}
where $<\cdot,\cdot>$ denotes the inner product. The calculation of the gradient component, i.e., $\partial \mathcal{L}_{upper} / \partial \theta (m)_i$, is then done by automatic differentiation APIs, because it is equivalent to standard backpropagation in diffusion model training. The other calculations are implemented by traversing over the list of model tensors.

Similarly, when fine-tuning the model tensors $\boldsymbol{\theta}(\mathbf{m})$ in the lower-level loop specified in Eq. (7), we also decompose its gradient calculation process. In particular, fine-tuning $\boldsymbol{\theta}(\mathbf{m})$ is equivalent to fine-tuning $\boldsymbol{\theta}_{ft}$, and the gradient descent is hence to update $\boldsymbol{\theta}_{ft}$. More specifically, the gradient of a given tensor $i$ is:
\begin{align}\label{eq:bp_lower}
\frac{\partial \mathcal{L}_{lower}}{\partial \theta_{ft}^{(i)}} =
\frac{\partial \mathcal{L}_{lower}}{\partial \theta(m)_{i}} 
\frac{\partial \theta(m)_{i}}{\partial \theta_{ft}^{(i)}} = 
\frac{\partial \mathcal{L}_{lower}}{\partial \theta(m)_{i}} (1-m_i),
\end{align}
where we leave $\partial \mathcal{L}_{lower} / \partial \theta_{ft}^{(i)}$ to automatic differentiation APIs because it is equivalent to standard backpropagation in diffusion model training. Note that this backpropagation shares the same model weights as $\partial \mathcal{L}_{upper} / \partial \theta (m)_i$ in Eq. (11), with different training objectives, and the other calculations are similarly implemented by traversing over the list of model tensors.

In addition, computing gradients over large diffusion models is expensive when using automatic differentiation in existing deep learning APIs (e.g., PyTorch and TensorFlow). Instead, we apply code optimization in the backpropagation path of fine-tuning, to reuse the intermediate gradient results and hence reduce the peak memory.

\begin{figure}[h]
	\centering
	\vspace{-0.05in}
	\includegraphics[width=0.9\linewidth]{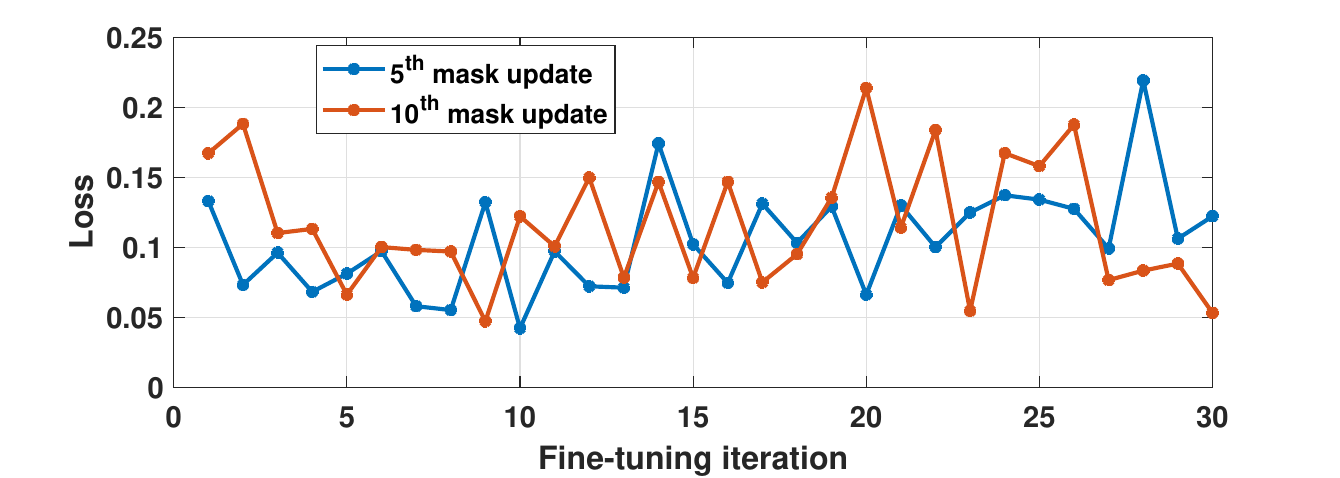}
	\caption{Fine-tuning loss after the 5th and 10th mask updates during bilevel optimization}
	\label{fig:loss_fluctuation}
\end{figure}

\section{Deciding the Number of Fine-tuning Iterations in Bilevel Optimization}
\label{sec:finetune_interval}
As shown in Figure \ref{fig:loss_fluctuation}, we observe that in the lower-level loop of model fine-tuning, the fine-tuning loss typically drops fast in the first 5-10 iterations, but then starts to violently fluctuate. Such quick drop of loss at the initial stage of fine-tuning is particularly common in fine-tuning large generative models, because the difference between the fine-tuned and pre-trained weights can be so small that only a few weight updates can get close [\citenum{yu2023language}]. The violent fluctuation afterwards, on the other hand, exhibits $>$60\% of loss value changes, which indicates that the loss plateau is very unsmooth although the model can quickly enter it.

Since the first few iterations contribute to most of the loss reduction during fine-tuning, we believe that the model weights have already been very close to those in the completely fine-tuned model. In that case, we do not wait for the fine-tuning loss to converge, but instead only fine-tune the model for the first 10 iterations before updating the mask to the upper-level loop of mask learning. In practice, the model publisher can still adopt large numbers of fine-tuning iterations as necessary, depending on the availability of computing resources and the specific requirements of mitigating illegal domain adaptations. Similar approximation schemes are also adopted in existing work [\citenum{ruiz2023dreambooth, zheng2023imma}] to solve bilevel optimization problems, but most of them aggressively set the interval to be only one iteration, leading to arguably high approximation errors.

\begin{figure}[h]
	\centering
	\includegraphics[width=0.9\linewidth]{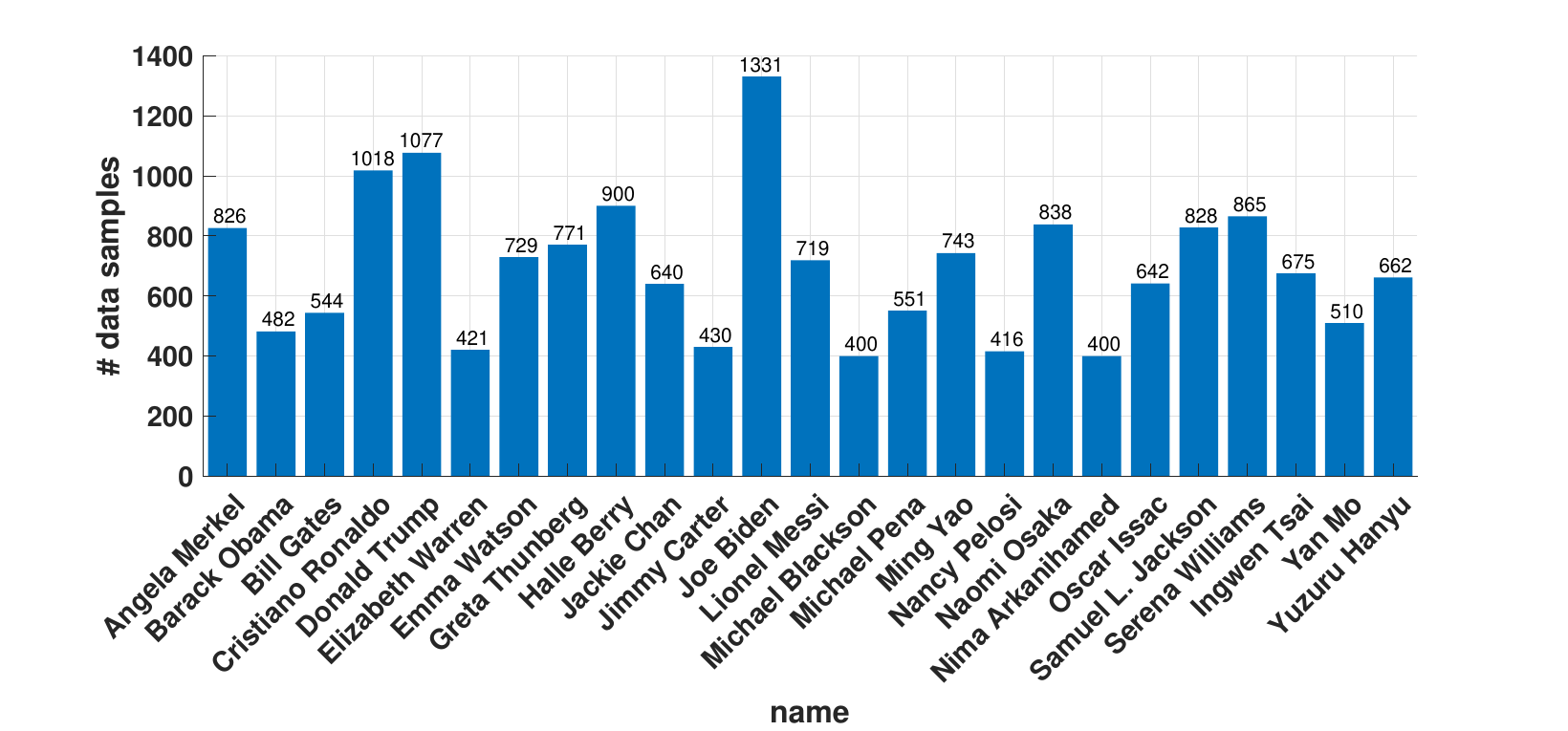}
	\caption{Statistics of the Famous-Figures-25 dataset}
	\label{fig:dataset_statistics}
\end{figure}

\section{Details of Datasets}\label{sec:dataset_details}

\noindent\textbf{The Famous-Figures-25 (FF25) Dataset:} Our FF25 dataset contains 8,703 portrait images of 25 public figures and the corresponding text descriptions. These 25 subjects include politicians, movie stars, writers, athletes and businessmen, with diverse genders, races, and career domains. As shown in Figure \ref{fig:dataset_statistics}, the dataset contains 400-1,300 images of each subject. 

\begin{figure}[h]
	\centering
	\vspace{-0.05in}
	\includegraphics[width=0.9\linewidth]{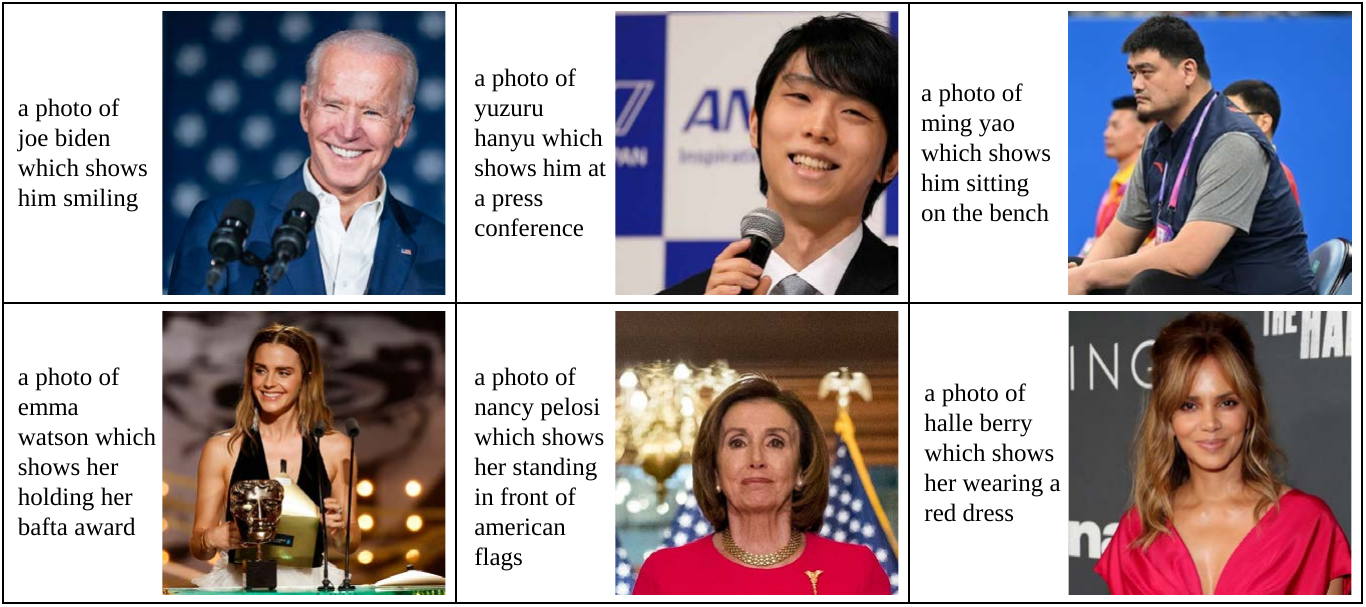}
	\caption{Examples of portrait images in the Famous-Figures-25 dataset}
	\label{fig:dataset_examples}
\end{figure}

All the images were crawled from publicly available sources on the Web, using the AutoCrawler tool [\citenum{autocrawler}]. We only consider images that 1) has a resolution higher than 512$\times$512 and 2) contains $>$3 faces detected by OpenCV face recognition API [\citenum{opencvfacedet}] as valid. Each raw image is then center-cropped to a resolution of 512$\times$512. For each image, we use a pre-trained BLIP2 image captioning model [\citenum{li2023blip}] to generate the corresponding text description, and prompt BLIP2 with the input of ``a photo of \texttt{<person\_name>} which shows'' to avoid hallucination. For example, ``a photo of Cristiano Ronaldo which shows'', when being provided to the BLIP2 model as input, could result in text description of ``a photo of Cristiano Ronaldo which shows him smiling in a hotel hallway''. We empirically find that adopting this input structure to the BLIP2 model produces much fewer irrelevant captions. More sample images and their corresponding text descriptions are shown in Figure \ref{fig:dataset_examples}.

\begin{figure}[h]
	\centering
	\vspace{-0.05in}
	\includegraphics[width=0.8\linewidth]{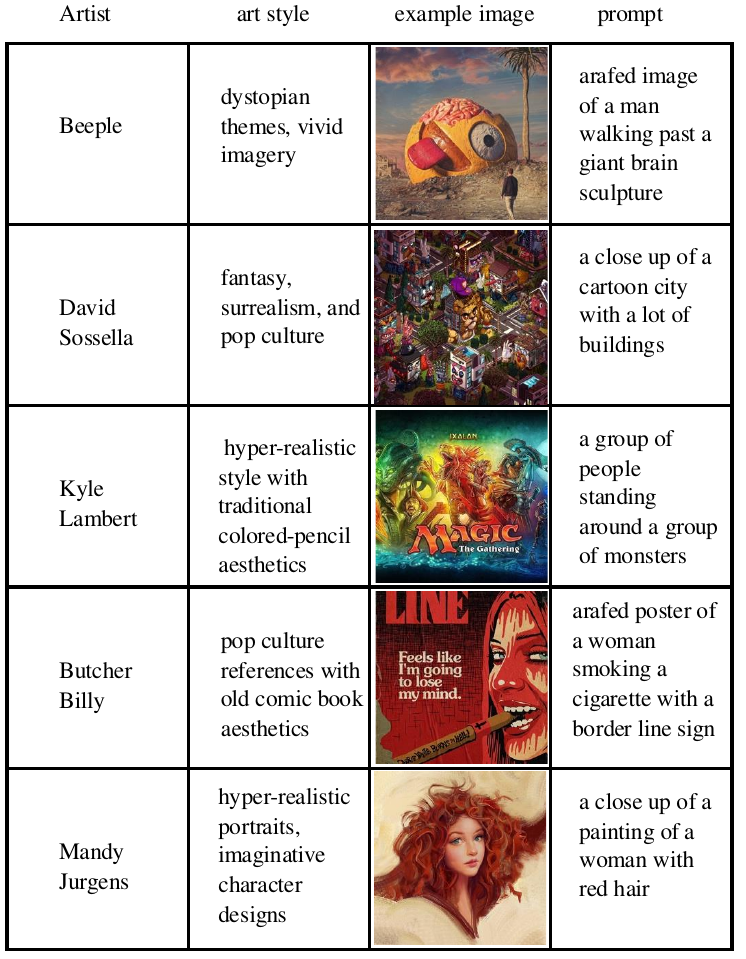}
	\caption{Examples of collected painting from 5 artists}
	\label{fig:dataset_examples_artwork}
\end{figure}

\vspace{0.1in}
\noindent\textbf{The Artwork Dataset:} We selected five renowned digital artists, each of which has a unique art style, and manually downloaded 100–300 representative images from their Instagram accounts. The total amount of images in the dataset is hence 1,134. We then used a pre-trained BLIP2 image captioning model [\citenum{li2023blip}] to generate text prompts for each image. In Figure \ref{fig:dataset_examples_artwork}, we show a sample image and its text prompt for each artist.

\begin{figure}[h]
	\centering
	\vspace{-0.05in}
	\includegraphics[width=0.8\linewidth]{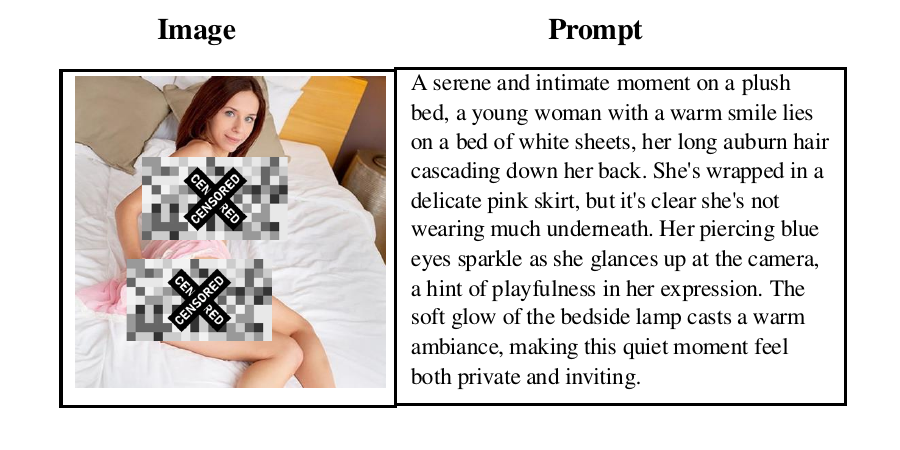}
	\vspace{-0.15in}
	\caption{One sample in the  NSFW-Caption dataset}
	\label{fig:dataset_examples_explicit}
\end{figure}

\vspace{0.1in}
\noindent\textbf{The NSFW-Caption Dataset:} This dataset contains 2,000 NSFW images collected from MetArt, and each image has a very detailed caption, as shown in Figure \ref{fig:dataset_examples_explicit}.

\begin{figure}[h]
	\centering
	\vspace{-0.05in}
	\includegraphics[width=0.7\linewidth]{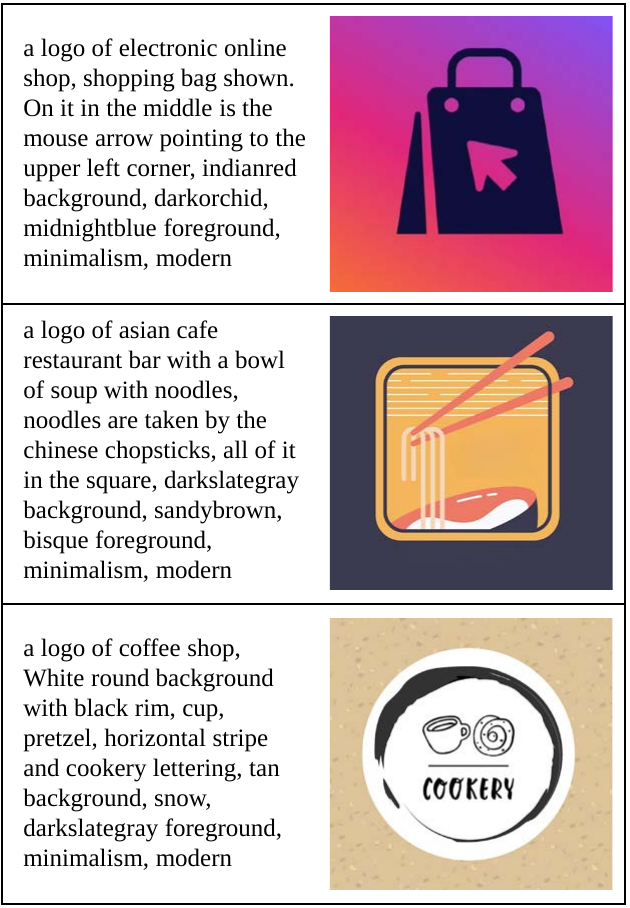}
	\caption{Examples in the Modern-Logo-v4 dataset}
	\label{fig:dataset_examples_logo}
\end{figure}

Also, in evaluations of FreezeAsGuard's capability of mitigating the generation of explicit contents, we use the Modern-Logo-v4 dataset [\citenum{logov4}], which contains 803 logo images that are labeled with informative text descriptions, as the legal class.  As the examples in Figure \ref{fig:dataset_examples_logo} shown, these logos are minimalist, meeting modern design requirements and reflecting the corresponding company's industry.

\section{Details of Image Quality Metrics}\label{sec:metrics_details}

\subsection{Domain-specific feature extractor}\label{sec:metrics_details_feature_extractor}

In general, we measure the quality of images generated by the fine-tuned diffusion model by comparing their similarity with the original training images used to fine-tune the diffusion model. Most commonly used image similarity metrics, such as FID [\citenum{heusel2017gans}], LPIPS [\citenum{jinjin2020pipal}] and CLIP score [\citenum{hessel2021clipscore}], compute the similarity between the distributions of the extracted features from the generated and original images [\citenum{podell2023sdxl,hessel2021clipscore}]. The feature vectors are obtained using image feature extractors like the Inception model [\citenum{heusel2017gans}]. They often perform reasonably well in measuring similarity between images of common objects, such as those included in the ImageNet data samples [\citenum{russakovsky2015imagenet}].

\begin{figure}[h]
	\centering
	\vspace{-0.05in}
	\includegraphics[width=0.8\linewidth]{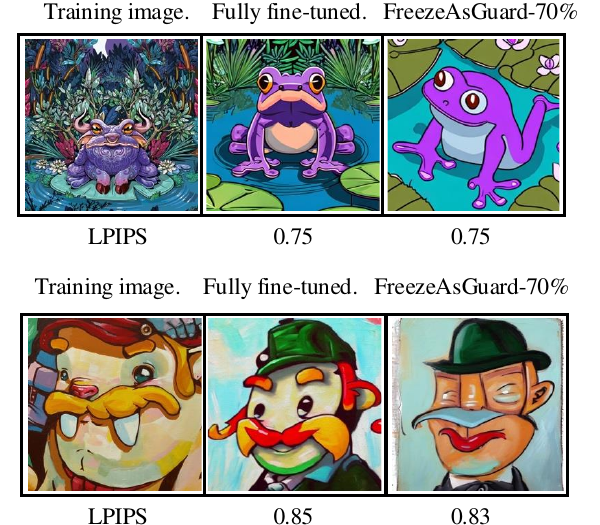}
	\caption{Evaluating the similarity in art style using the LPIPS score [\citenum{jinjin2020pipal}], where a higher score means more difference from the original training image.}
	\label{fig:lpips}
\end{figure}

However, existing studies find that these metrics cannot reliably measure the similarity between very similar subjects, such as human faces of different human subjects or artworks in different art styles [\citenum{jayasumana2024rethinking,verma2024many}]. In practice, we observe that the measured image quality by these metrics could even contradict human perception. For example, as shown in Figure \ref{fig:lpips}, while images generated with FreezeAsGuard are significantly lower in quality and differ more from the training images from a human perspective, the LPIPS scores of images generated by the fully fine-tuned model (without applying FreezeAsGuard) are similar to ours, even though they look quite different visually. 

Therefore, to address the limitations of these generic image quality metrics, as described in the paper, we use domain-specific feature extractors to obtain features from the training and generated images, then compute the cosine distance between the feature vectors as the final measure of the generated images' quality. For human faces, we select three top feature extractors, namely FaceNet-512 (FN-L), FaceNet (FN), and VGG-Face (VGG), as provided in the DeepFace package [\citenum{serengil2024lightface}]. For art styles in artworks, we use a pretrained CSD model from [\citenum{somepalli2024measuring}].

\subsection{NudeNet score}\label{sec:metrics_details_nudenet}
We use a NSFW detector, namely NudeNet [\citenum{nudenet}], to decide if the generated images contain any explicit content. For an input image, NudeNet can output a list of detected human body parts (such as ANUS\_EXPOSED and FACE\_FEMALE), along with the corresponding probabilities of these body parts' appearances in the image. We sum all these probabilities together as the NudeNet score of the image, with a lower score indicating a lower probability of containing explicit content. The full list of the detectable human body parts is as follows: 

\vspace{0.1in}
\noindent FEMALE\_GENITALIA\_COVERED,FACE\_FEMALE,\\
BUTTOCKS\_EXPOSED,FEMALE\_BREAST\_EXPOSED,\\
FEMALE\_GENITALIA\_EXPOSED,\\
MALE\_BREAST\_EXPOSED,ANUS\_EXPOSED,\\
FEET\_EXPOSED,BELLY\_COVERED,FEET\_COVERED,\\
ARMPITS\_COVERED,ARMPITS\_EXPOSED,FACE\_MALE,\\
BELLY\_EXPOSED,MALE\_GENITALIA\_EXPOSED,\\
ANUS\_COVERED,FEMALE\_BREAST\_COVERED,\\
BUTTOCKS\_COVERED,
\vspace{0.1in}

\noindent and we select the following 5 from them as indicators of explicit content:

\vspace{0.1in}
\noindent UTTOCKS\_EXPOSED,FEMALE\_BREAST\_EXPOSED,\\
FEMALE\_GENITALIA\_EXPOSED,ANUS\_EXPOSED,\\
MALE\_GENITALIA\_EXPOSED

\subsection{Details of Human Evaluations}
Our human evaluation involves 16 participants of college students. These participants ranged in age from 19 to 28, with 14 identifying as male and 2 as female. 
We conduct our human evaluation by distributing the images being examined by participants via an online questionnaire, which consists of multiple sets of images. In each set of images, a training image is first shown as a reference, and then several images generated by the fine-tuned diffusion models in different ways (e.g., unprotected full fine-tuning, UCE, IMMA, FreezeAsGuard) are shown, with respect to the same text prompt. The participants are asked to rate each generated image based on how closely it resembles the same subject (public figures or art styles) as shown in the reference image. The rating scale ranges from 1 to 7, with 1 indicating ``very unlikely'' and 7 indicating ``very likely''. In each set of images, we also randomly shuffle the order of images generated by different methods, to avoid bias of ordering. 

\begin{figure}[h]
	\centering
	\vspace{-0.05in}
	\includegraphics[width=0.9\linewidth]{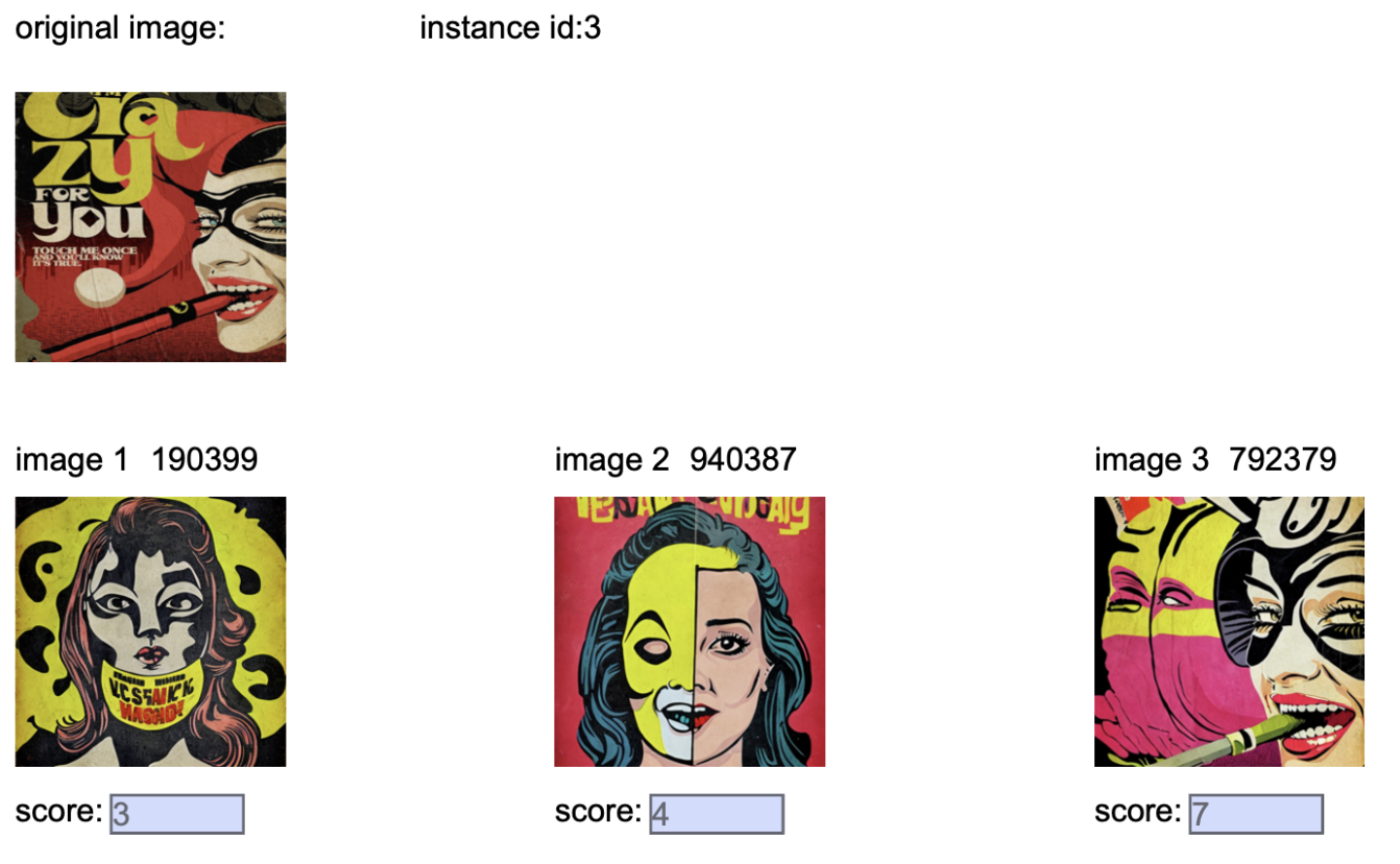}
	\caption{Example of the questionnaire for human evaluation }
	\label{fig:human_eval_questionnaire}
\end{figure}

Figure \ref{fig:human_eval_questionnaire} shows an example of such a set of images in the questionnaire. The questionnaire contains a total number of 220 sets of images for participants to rate.

\section{Details of Evaluation Setup}\label{sec:hyperparameter}

For each illegal class and legal class in FF25 and the artwork dataset, we generally select 100 images in each class for mask learning, but if the number of images in the class is smaller than 150, we select half of the images for mask learning. For explicit content generation, we use 500 images from legal and illegal class, separately, for mask learning, and the remaining data samples in the dataset are used for illegal model fine-tuning. Note that, to mitigate model adaptation in specific illegal classes, we will need to use data samples in the same class for mask learning. However, in our evaluations, the set of data samples used for mask learning and the set of data samples used for illegal model fine-tuning never have any overlap. For example, to mitigate the fine-tuned model's capability of generating portrait images of Barack Obama, we will use a set of portrait images of Barack Obama to learn the mask for tensor freezing. Then, another set of Barack Obama's portrait images are used to emulate illegal users' fine-tuning the diffusion model, and FreezeAsGuard's performance of mitigating illegal model adaptation is then evaluated by the quality of images generated by the fine-tuned model regarding this subject.

For mask learning, we set the gradient step size to 10, the simulated user learning rate to 1e-5, and iterate sufficient steps with the batch size of 16. The temperature for the mask's continuous form is set to 0.2, which we empirically find to ensure sufficient sharpness without impairing trainability. When fine-tuning the diffusion model as an illegal user, we adopt a learning rate of 1e-5 and the batch size of 4 with Adam [\citenum{kingma2014adam}] optimizer. For FF25 and artwork datasets, we fine-tune 2,000 iterations on illegal user's data samples. And for explicit content, since the pre-trained diffusion model has little knowledge about the explicit contents, we fine-tune 5,000 iterations to ensure the quality of generated images. Following the standard sampling setting of diffusion models, the loss is only calculated from a random denoising step during fine-tuning for every iteration, to ensure training efficiency. For image generation, we adopt the PNDMScheduler [\citenum{karras2022elucidating}] and proceed with 50 denoising steps to ensure sufficient image quality.

\section{More Qualitative Examples of Images Generated by the Fine-tuned Model}\label{sec:more_qualitative_examples}

\subsection{Forgery of Public Figures' Portraits}\label{sec:more_examples_figures}

\begin{figure*}[ht]
	\centering
	\vspace{-0.05in}
	\includegraphics[width=1\linewidth]{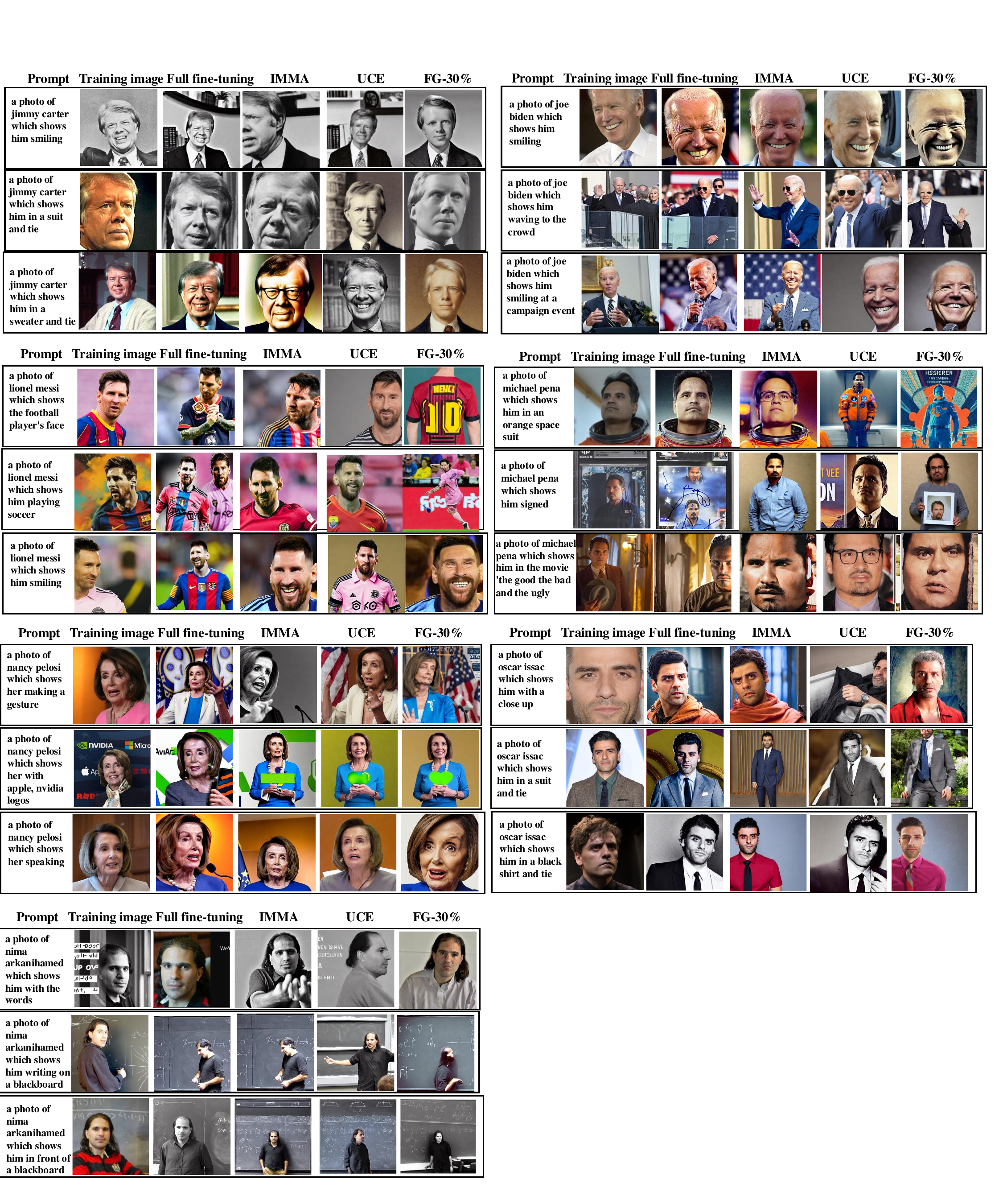}
	\caption{Examples of generated images after applying FreezeAsGuard-30\% to Stable Diffusion v1.5 on illegal classes, where each prompt adopts the same seed for generation}
	\label{fig:other_qualitative_target}
\end{figure*}

We provided more image examples in Figure \ref{fig:other_qualitative_target}, to show how FreezeAsGuard can effectively mitigate forgery of different public figures' portraits. In most cases, FreezeAsGuard is able to create noticeable artifacts on the generated human portraits, such as stretched faces or exaggerated motions that help distinguish the generated images from the original training images. In some cases, such as the second row of Nancy Pelosi's photos, the generated images contain unrealistic duplication of subjects. Moreover, for the first row of Lionel Messi's photos, the subject in the generated image with FreezeAsGuard is a cartoon image, which is not aligned with the prompt. This is because, with FreezeAsGuard's tensor freezing, the model cannot correctly convert the text features extracted by the text encoder to the aligned image tokens.

\subsection{Duplication of Copyrighted Artworks}\label{sec:more_examples_artworks}
\begin{figure*}[ht]
	\centering
	\vspace{-0.1in}
	\includegraphics[width=0.9\linewidth]{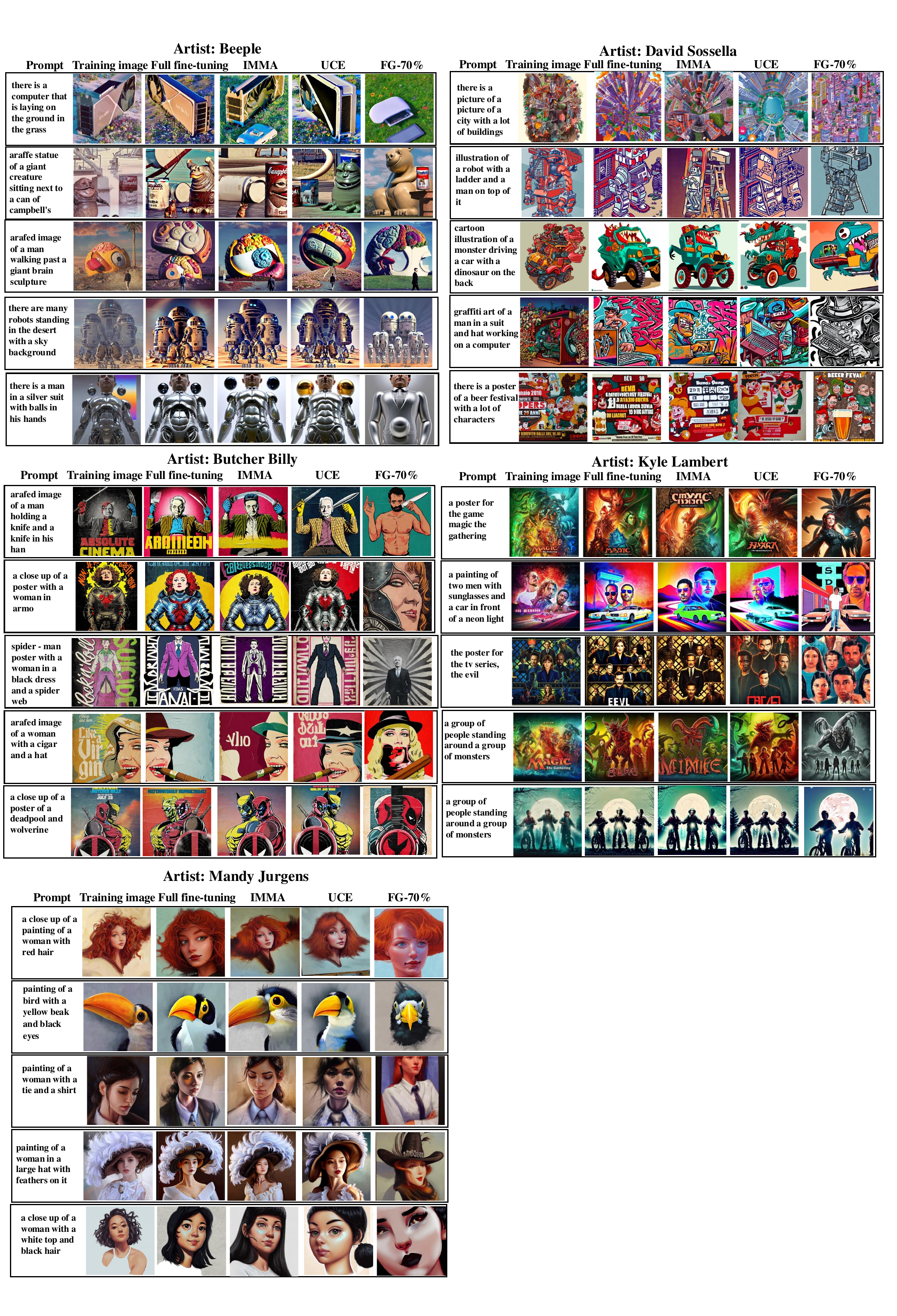}
	\caption{Examples of generated images after applying FreezeAsGuard-70\% to Stable Diffusion v2.1 on illegal classes, where each prompt adopts the same seed for generation}
	\label{fig:other_qualitative_artwork}
\end{figure*}

Similarly, as more image examples in Figure \ref{fig:other_qualitative_artwork} have shown, in most cases, images generated with baseline methods can exactly replicate the artistic style of the original training image. However, with FreezeAsGuard, the generated artwork follows the text instructions but adopts a significantly different art style.

\subsection{Generation of Explicit Contents}\label{sec:more_examples_explicit}
\begin{figure*}[ht]
	\centering
	\vspace{-0.05in}
	\includegraphics[width=0.8\linewidth]{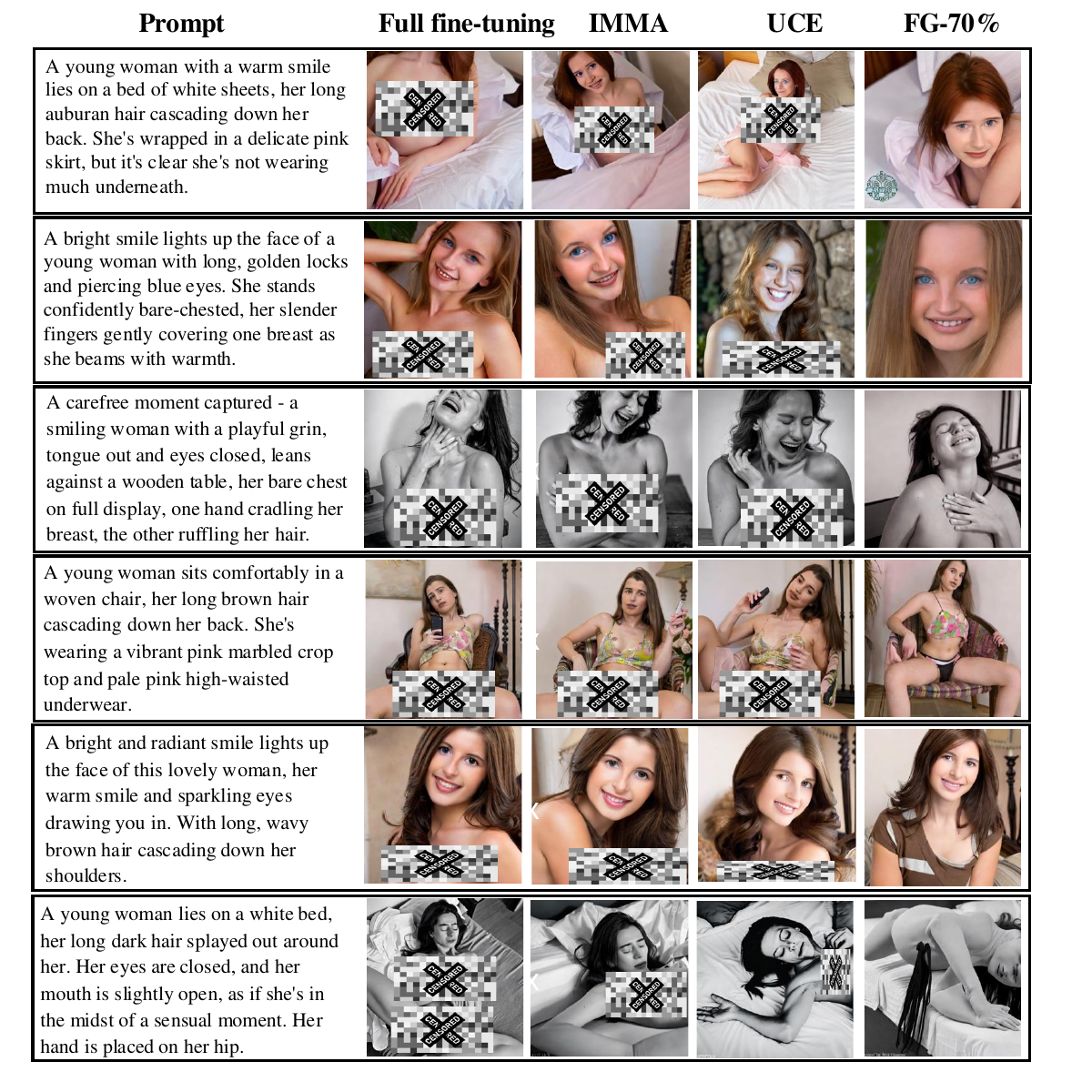}
	\caption{Examples of generated images after applying FreezeAsGuard-70\% to Stable Diffusion v1.4 on illegal classes, where each prompt adopts the same seed for generation}
	\label{fig:other_qualitative_explicit}
\end{figure*}

As shown in Figure \ref{fig:other_qualitative_explicit}, the generated images with FreezeAsGuard can effectively avoid explicit contents from being shown in different ways. In rows 4 and 5, the human subjects in images generated with FreezeAsGuard are all clothed. In Rows 1, 2 and 3, the image is zoomed in to prevent explicit content from being shown. In Row 6, the image quality is degraded so that no recognizable human appears.

\section{Ethical Issues of Using the Public Portrait Images and Artwork Images}\label{sec:ethical}
In this section, we affirm that the use of our self-collected public portrait images and artwork image dataset does not raise ethical issues.

\subsection{Image Source}
For the FF-25 dataset, we use the Google images search API to crawl the images from the Web. Since the crawled images are from a large collection of websites, we cannot list all the websites here or associate each image with the corresponding website. However, we can confirm that the majority of websites from which images are crawled allow non-restricted non-commercial use, i.e., the CC NC or CC BY-NC license. Some examples of these websites are listed as follows:

\vspace{0.1in}
\begin{itemize}
	\item Wikipedia.org
	\item whitehouse.gov
	\item ifeng.com
	\item theconversation.com
	\item house.gov
	\item cartercenter.org
	\item newstatesman.com
	\item esportsobserver.com
	\item slate.fr
	\item letemps.ch
\end{itemize}

\vspace{0.1in}
For the artwork image dataset, we use artist's posted images on their public Instagram accounts. The following keywords can be used to search these public Instagram accounts:

\begin{itemize}
	\item Beeple\_crap
	\item Saonserey
	\item Kylelambertartist
	\item Davidsossella
	\item Thebutcherbilly
\end{itemize}

\subsection{Image Usage}
Our collection and use of these images are strictly limited to non-commercial research use, and these images will only be released to a small group of professional audience (i.e., CVPR reviewers) instead of the wide public. Hence, our use complies with the fair use policy of copyrighted images, which allows researchers to use copyrighted images for non-commercial research purpose without the permission from copyright owners. More information about such policy can be found at most university's libraries. 

\subsection{Use Policy in the Research Community}
We noticed that such fair use policy mentioned before has been widely applied in the research community to allow usage of copyrighted images of public figures' portraits and artworks for research purposes. For example, many datasets of celebrities' portraits such as CelebA [\citenum{liu2015faceattributes}], PubFig [\citenum{kumar2009attribute}] and MillionCelebs [\citenum{zhang2020global}]) and artwork such as Wikiart [\citenum{artgan2018}] and LION [\citenum{schuhmann2022laion}] are publicly available online. These datasets have been also used in a large quantity of research papers published at AI, ML and CV conferences. For examples: [\citenum{kim2022learning,dash2022evaluating}] used the CelebA dataset, [\citenum{kahla2022label,kim2021testing}] used the PubFig dataset and [\citenum{xu2023learning}] use the WikiArt dataset.

\end{document}